\documentclass{article}

\usepackage{amsmath,amsfonts,bm}

\def\eqref#1{equation~\ref{#1}}
\def\1{\bm{1}}

\DeclareMathAlphabet{\mathsfit}{\encodingdefault}{\sfdefault}{m}{sl}
\SetMathAlphabet{\mathsfit}{bold}{\encodingdefault}{\sfdefault}{bx}{n}

\def\gA{{\mathcal{A}}}
\def\gB{{\mathcal{B}}}

\def\gD{{\mathcal{D}}}

\def\gN{{\mathcal{N}}}

\def\gP{{\mathcal{P}}}

\def\gS{{\mathcal{S}}}
\def\gT{{\mathcal{T}}}

\def\sN{{\mathbb{N}}}

\def\sR{{\mathbb{R}}}

\newcommand{\E}{\mathbb{E}}

\usepackage{microtype}
\usepackage{graphicx}
\usepackage{subcaption}
\usepackage{booktabs} % for professional tables

\usepackage{hyperref}

\usepackage[accepted]{icml2026}

\usepackage{amsmath}
\usepackage{amssymb}
\usepackage{mathtools}
\usepackage{amsthm}

\usepackage[capitalize,noabbrev]{cleveref}

\theoremstyle{plain}
\newtheorem{theorem}{Theorem}[section]
\newtheorem{proposition}[theorem]{Proposition}

\theoremstyle{definition}

\theoremstyle{remark}

\usepackage[textsize=tiny]{todonotes}

\icmltitlerunning{Offline Reinforcement Learning with Universal Horizon Models}

\begin{document}

\twocolumn[
  \icmltitle{Offline Reinforcement Learning with Universal Horizon Models}



  \icmlsetsymbol{equal}{*}

  \begin{icmlauthorlist}
    \icmlauthor{Hojun Chung}{equal,ipai}
    \icmlauthor{Junseo Lee}{equal,ipai}
    \icmlauthor{Songhwai Oh}{ipai,ece}
  \end{icmlauthorlist}

  \icmlaffiliation{ipai}{Interdisciplinary Program in Artificial Intelligence and ASRI, Seoul National University}
  \icmlaffiliation{ece}{Department of Electrical and Computer Engineering, Seoul National University}

  \icmlcorrespondingauthor{Songhwai Oh}{songhwai@snu.ac.kr}

  \icmlkeywords{Offline reinforcement learning, Model-based reinforcement learning, World models}

  \vskip 0.3in
]



\printAffiliationsAndNotice{\icmlEqualContribution}

\begin{abstract}
Model-based reinforcement learning (RL) offers a compelling approach to offline RL by enabling value learning on imagined on-policy trajectories. 
However, it often suffers from compounding errors due to repeated model inference on self-generated states.
While geometric horizon models (GHM) alleviate this issue through direct prediction over a discounted infinite-horizon future, they remain challenged in accurately modeling distant future states. To this end, we introduce universal horizon models (UHM), a generalization of GHM that directly predicts future states under arbitrary horizons. Leveraging this flexibility, we propose a scalable value learning method that employs a winsorized horizon distribution to stabilize training by capping excessively large horizons. 
Experimental results on 100 challenging OGBench tasks demonstrate that
the proposed method outperforms competitive baselines, particularly on tasks with highly suboptimal datasets and those requiring long-horizon reasoning. Project page: \href{https://rllab-snu.github.io/projects/UHM/}{https://rllab-snu.github.io/projects/UHM/}

\end{abstract}

 % over a discounted infinite-horizon future

\section{Introduction}
Offline reinforcement learning \cite{offlinerl_levine, prudencio2023survey} provides a promising way to learn effective policies without online exploration, enabling the utilization of pre-collected datasets. However, its scalability is often restricted in tasks that require long-horizon reasoning due to the bias accumulation in temporal difference (TD) learning \cite{latent_rosete, horizon_park}. 
Recent works reveal the potential to solve such complex tasks using $n$-step TD, which utilizes $n$-step returns for Bellman backups \cite{rl_sutton} to reduce the bias in the TD target \cite{horizon_park, transitive_park}. Despite the benefit, model-free approaches should rely on trajectories in the dataset since they have no ability to imagine on-policy rollouts. This distributional mismatch ruins a theoretical foundation of TD learning, and can lead to inaccurate value estimation \cite{multistep_de, multistep_hernandez}.

Model-based reinforcement learning (MBRL) can address this issue with model-based value expansion, i.e., on-policy value learning on synthetic trajectories \cite{mve_feinberg, dreamerv4_hafner}. 
However, generating such trajectories with single-step dynamics models requires repeated model inference on self-generated states, where keeping the dynamics error small is challenging.
While geometric horizon models (GHM) mitigate this issue by directly predicting future states, they lack the ability to predict future states at a specified timestep. This limitation forces GHM to model long-horizon tails of the geometric distribution, which is inherently difficult to learn accurately.

% While geometric horizon models (GHM) mitigate this issue by directly predicting future states given real state-action pairs
% , they lack the ability to predict future states at a specified timestep. This limitation forces GHM to model long-horizon tails of the geometric distribution, which is inherently difficult to learn accurately.

\begin{figure*}[t!]
  \centering
  \includegraphics[width=1.0\linewidth]{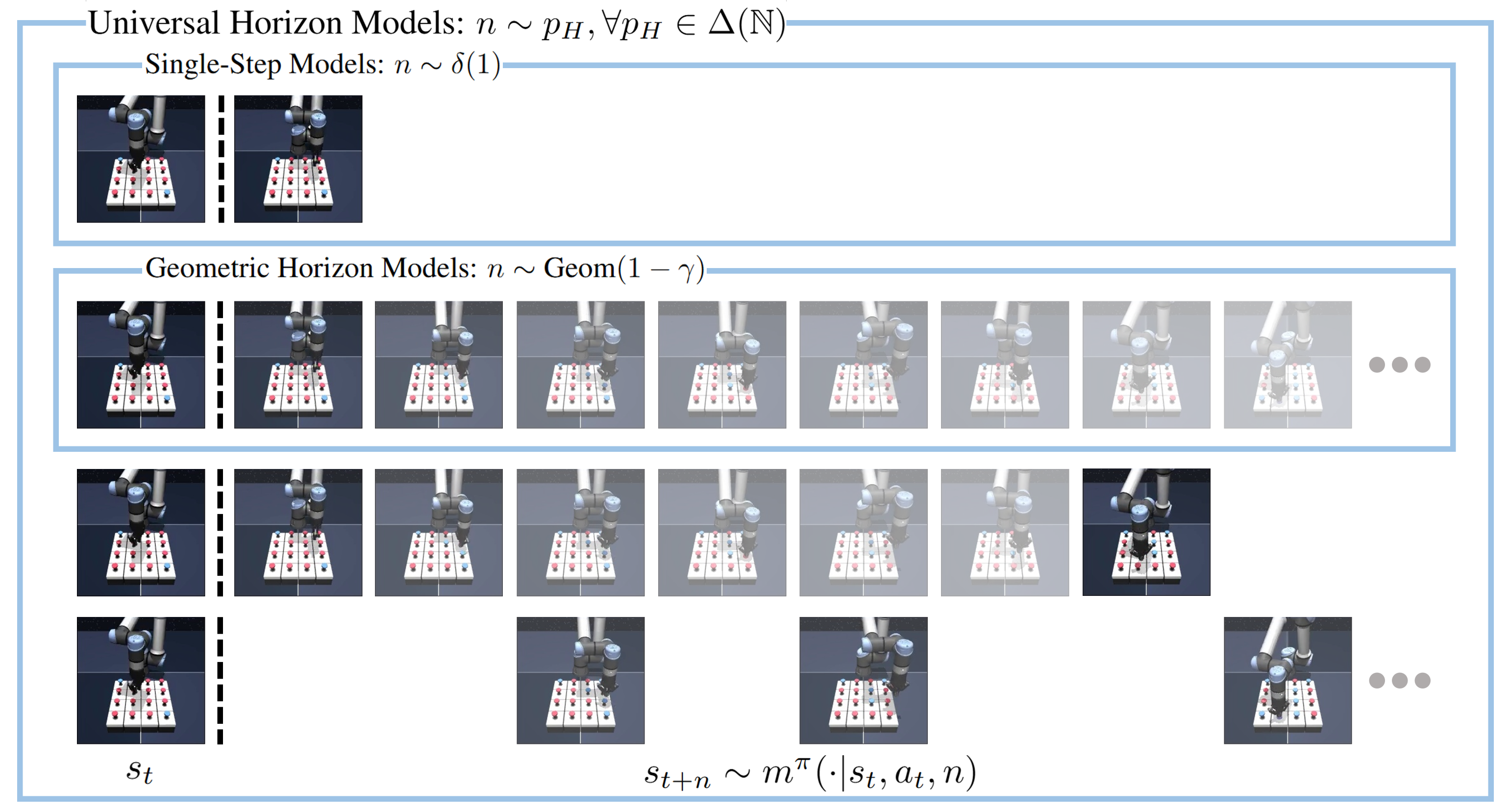} 
\caption{Universal horizon model is a future predictive model that directly samples states from the $n$-step future state distribution of the policy for any given horizon $n$. Since it allows $n$ to be sampled from arbitrary distributions, UHM can be seen as a general framework that includes single-step models and geometric horizon models.}
\label{fig:uhm}
\end{figure*}

In this work, we focus on developing a scalable model-based value learning method for offline RL. We first propose universal horizon models (UHM), which can sample states directly from $n$-step future state distributions for any given $n$. As illustrated in Figure \ref{fig:uhm}, UHM generalizes both GHM and single-step models, as it allows $n$ to be sampled from arbitrary horizon distributions.
This generalization enables a TD learning method that allows flexible control over which future horizons to focus on. Building on this framework, we present a value learning method using winsorized future horizons to stabilize the learning process by capping the maximum value of $n$. 
Through extensive experiments, we demonstrate that the proposed method outperforms baselines across 100 challenging tasks in  OGBench \cite{ogbench_park2025}, including tasks that provide highly suboptimal datasets or require long-horizon reasoning. In summary, the main contributions of the paper are as follows:
\begin{itemize}
    \item We propose universal horizon models (UHM), which generalize geometric horizon models by directly sampling $n$-step future states for any given $n$.
    \item We introduce a robust and scalable model-based value expansion method using UHM.
    \item We demonstrate that the proposed method achieves a 14\% higher average success rate than the strongest baseline across 100 challenging OGBench tasks.

\end{itemize}

\section{Related Work}
\textbf{Offline RL}~\cite{offlinerl_levine} studies the problem of learning policies from static datasets without online interaction. A wide range of offline RL algorithms are built upon temporal difference (TD) learning, which updates the critic using one-step lookahead value estimation. To adapt TD learning to offline settings, prior work has proposed uncertainty-aware methods~\cite{q_ensemble_an, q_ensemble_gu, uncertainty_wu}, conservative updates~\cite{cql-kumar, dual_sikchi}, and in-sample maximization techniques~\cite{iql_kostrikov, xql_garg, insample_xu}. Policies are then learned from the critic via advantage-weighted regression~\cite{weighted_peters, awr_peng}, behavior-regularized policy gradients~\cite{td3bc_fujimoto, rebrac_tarasov, fql_park}, or using generative modeling~\cite{sfbc_chen, idql_hansen}. Recently, horizon reduction techniques have demonstrated strong performance on complex tasks by training hierarchical policies~\cite{latent_rosete, 0parrot_singh}, leveraging action chunking~\cite{sequence_seo, chunking_rl_li, dqc_li}, or adopting multi-step TD methods~\cite{horizon_park, transitive_park}.

\textbf{Offline MBRL} has been studied with its potential to generate on-policy trajectories without environment interaction. Despite its success in online settings \cite{mbpo_janner, td-mpc2_hansen, dreamerv4_hafner}, dynamics models learned from static datasets have inevitable errors \cite{model_talvitie}, making it challenging to directly apply it to offline settings. To resolve this issue, previous works consider uncertainty penalization \cite{kidambi2020morel, mopo_yu, mobile_sun}, or conservative critic learning \cite{leq_park}. Another branch of MBRL leverages generative models to learn a distribution of trajectories \cite{decision_transformer_janner, diffuser_janner, dtamp_hong}, and guide them to produce trajectories with higher expected returns \cite{ajay2022conditional, policy_guided_diffusion_jackson, cheng2024scaling}. Meanwhile, geometric horizon models \cite{gamma_janner, ghm_thakoor} have emerged as an approach for directly predicting states from the discounted future. Combined with flow-matching \cite{flow_matching_lipman}, it shows superior performance on both future prediction and value estimation \cite{tdflow_farebrother, infom_zheng}. Our work is also closely related to the prior works that employ dynamics models to predict future beyond the next timestep \cite{zhang2023leveraging, lin2024any, mac_park}. 
These methods train models to predict $\text{Pr}(s_{t+n}|s_t, a_t, \dots, a_{t+n-1})$, which is a future determined by fixed-length action chunks. In contrast, our approach learns $\text{Pr}(s_{t+n}|s_t,a_t,\pi)$ for an arbitrary horizon $n$, which is a future state distribution induced by the policy. This distinction allows us to generate horizon-flexible future states without repeated inference, thereby reducing computational overhead and avoiding recursive conditioning on self-generated states.

\section{Preliminaries}
Reinforcement learning (RL) can be formulated as an infinite horizon Markov decision process $(\mathcal{S}, \mathcal{A}, R, \gamma, \gP, \rho)$, where $\mathcal{S}$ is a state space, $\mathcal{A}$ is an action space, $R:\mathcal{S} \times \mathcal{A} \rightarrow \mathbb{R}$ is a reward function, $\gamma \in (0,1)$ is a discount factor, $\gP: \mathcal{S} \times \mathcal{A} \rightarrow \Delta(\mathcal{S})$ is a transition probability, and $\rho
\in\Delta(\mathcal{S})$ is an initial state distribution. Offline RL aims to find a policy $\pi:\mathcal{S}\rightarrow\Delta(\mathcal{A})$ that maximizes a cumulative discounted return $\mathbb{E}\left[\sum_{k=0}^{\infty}\gamma^k R(s_k, a_k) \mid s_0\sim\rho(\cdot), a_k\sim\pi(\cdot|s_k) \right]$ from the offline dataset $\mathcal{D} = \{(s_k^m,a_k^m, s_{k+1}^m)_{k=0}^{H_m}\}_{m \in \{1,2, \dots M\}}$.

\subsection{Temporal Difference Learning for Offline RL}
Value learning in offline RL is typically based on temporal difference (TD) learning. Given a transition tuple $(s,a,s')\sim\gD$, a critic $Q(s,a)$ is trained to approximate $Q^\pi(s,a) = \mathbb{E}\left[\sum_{k=0}^{\infty}\gamma^k r_k \vert s_0=s, a_0=a, \pi\right]$ by regressing towards its bootstrapping target $R(s,a) + \gamma Q(s',a')$, where $a' \sim \pi(\cdot|s')$.
Besides its convergence guarantee, TD learning often suffers from bias accumulation, which originates from inaccurate value estimates in the bootstrapping target \cite{rl_sutton}. 

A common approach to reduce the bias is $n$-step TD, which calculates the bootstrapping target based on the trajectory $(s_{k}, a_{k}, s_{k+1})_{k=0}^{n-1}$ whose length is $n \in \sN$:
\begin{equation}
\label{eq_td_n}
    G^{(n)} = \sum_{k=0}^{n-1} \gamma^kR(s_{k}, a_{k}) + \gamma^nQ(s_{n}, a_{n}),
\end{equation}
where $a_n \sim \pi(\cdot|s_n)$. TD($\lambda$) generalizes $n$-step TD by using a weighted average of bootstrapping targets from various $n \in \sN$ with decay parameter $\lambda \in \left[0,1\right]$:
\begin{equation}
\label{eq_td_lambda}
    G^{\lambda} = (1-\lambda)\sum_{n=1}^\infty \lambda^{n-1}G^{(n)}.
\end{equation} 
Both $n$-step TD and TD($\lambda$) are common techniques to boost the performance in online RL \cite{gae_schulman, hessel2018rainbow}, and recent works have shown that using $n$-step returns is also helpful to solve complex, long-horizon reasoning tasks in offline settings \cite{horizon_park, transitive_park}. However, model-free approaches rely on behavior trajectories in offline settings, making it unclear which policy the learned value function estimates \cite{multistep_de, multistep_hernandez}.

Model-based RL can resolve this issue by performing model-based value expansion (MVE)  \cite{mve_feinberg}, i.e., TD learning on synthetic on-policy rollouts. In contrast to the success in online RL \cite{dreamerv4_hafner}, dynamics models learned from static datasets are not globally accurate, exacerbating the compounding error problem induced by repeated inference of the dynamics model and policy. As a result, offline MBRL methods are often limited to restricted forms, such as conservative critic updates or short-horizon imagination~\cite{kidambi2020morel, mopo_yu, mobile_sun, leq_park}.

\subsection{Geometric Horizon Models}
Geometric horizon model (GHM) \cite{gamma_janner} is defined as a generative model of normalized successor measures \cite{successor_dayan, successor_blier}:
{\small
\begin{equation}
\label{eq_ghm_definition}
m^\pi(x|s,a) = (1 - \tilde{\gamma})\sum_{k=0}^\infty \gamma^k \text{Pr}(s_{k+1}=x \mid s_0=s, a_0=a, \pi),
\end{equation}}
which is a $\tilde{\gamma}$-discounted distribution of future states when deploying the policy $\pi$, where $\tilde{\gamma} \in (0, \gamma]$. The GHM can be learned from off-policy transitions via regression towards its bootstrapping target:
\begin{align}
\label{eq_ghm_bootstrap_target}
m^\pi(x|s,a) \leftarrow \, &(1 - \tilde{\gamma})\gP(x|s,a) \nonumber \\
&+ \tilde{\gamma} \mathbb{E}_{\substack{s' \sim \gP(\cdot|s,a) \\ a' \sim \pi(\cdot|s')}} \left[m^\pi(x | s', a')\right],
\end{align}
which is a contraction mapping and applicable for various generative modeling methods \cite{ghm_thakoor, tdflow_farebrother}.

% Since GHM enables direct sampling from the discounted future state distribution, 
% we can estimate the value as $Q^\pi(s,a) = R(s,a) + \frac{\gamma}{1-\gamma}\mathbb{E}_{s_e\sim m^\pi(\cdot\vert s,a), a_e\sim\pi(\cdot|s_e)}\left[R(s_e, a_e)\right]$. 
Previous works propose $\gamma$-MVE \cite{gamma_janner, ghm_thakoor} to obtain bootstrapping value target with future states sampled by GHM. The corresponding value target can be simplified as follows:
\begin{align}
\label{eq_gamma_mve}
Q_{\gamma\text{-MVE}} = R(s,a) + \gamma&\mathop{\mathbb{E}}_{\substack{s_e \sim m^\pi(\cdot\vert s,a) \\ a_e \sim \pi(\cdot|s_e)}} \bigg[\frac{1}{1-\tilde{\gamma}}R(s_e, a_e) \nonumber \\
&+ \frac{\gamma-\tilde{\gamma}}{1-\tilde{\gamma}}Q(s_e,a_e)\bigg].
\end{align}
% Detailed derivation is provided in Appendix \ref{subsec:td_lambda and gamma_mve}.
We found that $Q_{\gamma\text{-MVE}}$ (\ref{eq_gamma_mve}) is equal to the expectation of TD($\lambda$) target (\ref{eq_td_lambda}) on on-policy trajectories when $\lambda=\tilde{\gamma}/\gamma$, which means $\gamma$-MVE is a method to conduct TD($\lambda$) using GHM. Detailed derivation is provided in Appendix \ref{subsec:td_lambda and gamma_mve}.

\section{Proposed Methods}
In this section, we introduce a \emph{universal horizon model} (UHM), a future predictive model that generalizes geometric horizon models and single-step dynamics models. UHM directly samples $n$-step future states, and allows the horizon $n$ to be sampled from arbitrary distributions. This flexibility enables a general form of value estimation that recovers both $n$-step TD and TD($\lambda$). Based on this framework, we propose a scalable offline RL algorithm that stabilizes the learning process by capping excessively large horizons.

\subsection{Universal Horizon Models}
While GHMs avoid repeated inference on self-generated intermediate states by directly sampling from the successor measure (\ref{eq_ghm_definition}), they do not reveal how many steps into the future a sampled state lies. Moreover, their horizons are restricted to a single geometric distribution. As a result, they require accurate modeling of the long-horizon tail, which is inherently difficult to learn. To generalize beyond the implicit geometric horizon, we define a universal horizon model as a generative model of $n$-step transition measure:
\begin{equation}
\label{eq_uhm_measure}
  m^\pi(x \mid s,a,n) = \Pr(s_n = x \mid s_0=s, a_0=a, \pi),
\end{equation}
where the horizon $n$ can be sampled from any distribution $p_H$. The resulting marginal measure  
\begin{align}
  &m^{\pi,p_H}(x \mid s,a) 
  = \sum_{k\ge 1} p_H(k) m^\pi(x \mid s,a,k), \\
  &= \sum_{k\ge 1} p_H(k) \Pr(s_k = x \mid s_0=s, a_0=a, \pi),
\end{align}
represents a horizon-weighted visitation measure over future states. Since it recovers normalized successor measure when $n \sim \mathrm{Geom}(1-\gamma)$, GHM can be seen as a special case of UHM. Furthermore, we can also represent single-step dynamics models with $n \sim \delta(1)$. Hence, UHM provides a unified framework for future prediction that subsumes both GHM and single-step dynamics models, as illustrated in Figure~\ref{fig:uhm}.

UHM can be learned from off-policy transitions via bootstrapping:
\begin{align}
&m^\pi(x \mid s,a,1)=\gP(x \mid s,a), \\
&m^\pi(x \mid s,a,n+1) \nonumber\\
\label{eq_uhm_bootstrap}
&= \mathbb{E}_{s'\sim\gP(\cdot \mid s,a), a'\sim\pi(\cdot \mid s')}
\left[m^\pi(x \mid s',a',n)\right],
\end{align}
analogous to the learning objective of GHM (\ref{eq_ghm_bootstrap_target}). Specifically, after sampling $n\sim p_H(\cdot)$ and $a' \sim \pi(\cdot|s')$, we train UHM so that its prediction for the $(n+1)$-step future state of $(s,a)$ matches a bootstrapped sample $s_e \sim m^\pi (\cdot|s',a',n)$. Since the multi-step target is defined recursively through bootstrapping, learning the one-step case correctly provides the basis for learning the $n > 1$ future state distributions.
% This recursion inductively reflects the transition dynamics $\gP$ and the policy $\pi$. 
Note that $m^\pi(x \mid s,a,n)$ is defined for each $n$, regardless of the horizon distribution $p_H$. Consequently, different $p_H$ can be used throughout training or at inference time without altering the definition of $m^\pi$.
% Note that the policy $\pi$ may differ from the dataset behavior, unlike generative models that learn the distribution of offline trajectories.

\subsection{Critic Target Estimation}
Based on the flexibility of UHM to represent and sample from arbitrary future distributions, we propose a generalized temporal difference learning framework.
\begin{proposition}
\label{proposition_uhm_convergence}
    For any sub-probability measure $\nu$ over $\sN$, consider the $\nu$-Bellman operator $\gT^\nu$ defined as
    \begin{align}
    \label{eq_nu_bellman_target}
        &\gT^\nu Q(s,a)\coloneqq\E\Bigl[R(s,a)+\gamma\sum_{k\ge 1}\bigl[
        \xi^\nu(k)R(s_{k}, a_{k}) \nonumber\\
        &\quad\quad\quad+ \nu(k)Q(s_{k},a_{k})\bigr]
        \Bigm\vert s_0=s, a_0=a, \pi \Bigr],
    \end{align}
    where 
    \begin{equation}
        \label{eq_xi}
        \xi^\nu(k) = \gamma^{k-1}-\sum_{\kappa=0}^{k-1}\left[
        \gamma^{\kappa}\nu(k-\kappa)\right].
    \end{equation}
    The iterative sequence $Q_{n+1}=\gT^\nu Q_n$ from any bounded real function $Q_0\colon\gS\times\gA\to\sR$ converges to $Q^\pi(s,a)$.
\end{proposition}
A proof is provided in Appendix \ref{subsec:proposition proof}. The TD learning with $\nu$-Bellman operator recovers standard critic updates as special cases: choosing $\nu(k)=\gamma^{n-1}\mathbf{1}[k=n]$ yields $n$-step TD, while choosing $\nu(k)=(1-\lambda)(\lambda\gamma)^{k-1}$ represents TD($\lambda$). To enable TD learning with the proposed framework (\ref{eq_nu_bellman_target}) while avoiding repeated inference, we utilize UHM as it can sample states from the $n$-step future state distribution for any given $n$. Specifically, UHM allows one-sample estimation $G^\nu$ of the $\nu$-Bellman backup target $\gT^\nu Q(s,a)$ for any sub-probability measure $\nu$ as follows:
% UHM enables estimating $\gT^\nu Q(s,a)$ for arbitrary sub-probability measure $\nu$ since it can sample from the $n$-step future state distribution for any given $n$. 
% Given an offline transition sample $(s,a,r)$, UHM allows one-sample estimation $G^\nu$ of the $\nu$-Bellman backup target $\gT^\nu Q(s,a)$ (\ref{eq_nu_bellman_target}) by sampling a future state $s_e$ conditioned on a random horizon $n$ that follows a distribution $p_H$:
\begin{align}
    n &\sim p_H, \\
    s_e &\sim m^\pi(\cdot|s,a,n), a_e \sim \pi(\cdot|s_e), \\
    \label{eq_uhm_td_target}
    G^\nu &= r + \gamma \left\{ w_\xi R(s_e, a_e)+ w_\nu Q(s_e, a_e) \right\},
\end{align}
where $p_H$ is a horizon distribution over $\mathbb{N}$ satisfying 
% $\mathrm{supp}(\nu) \cup \mathrm{supp}(\xi^\nu) \subseteq \mathrm{supp}(p_H)$, i.e.,
$p_H(k) > 0$ whenever $\nu(k) \neq 0$ or $\xi^\nu(k) \neq 0$.
The weights $w_\xi=\frac{\xi^\nu}{p_H}$ and $w_\nu=\frac{\nu}{p_H}$ are importance ratios that correct for the discrepancy between $\nu, \xi^\nu$ and the sampling distribution $p_H$. 

% Specifically, assigning more weight to the bootstrapped critic terms reduces variance by averaging over value estimates, but introduces bias due to approximation errors. 
% Conversely, assigning less weight to the critic makes the target rely more heavily on individual trajectory samples, which reduces bias but increases variance of the estimated target. 

% As in TD($\lambda$), the reward/critic weights across horizons $k$ induces a trade-off: 
% higher weight on the critic makes the target more sensitive to critic errors, while higher weight on large $k$
% makes it more sensitive to model errors.
% Since UHM directly samples from the $n$-step future state distribution for any given $n$, it enables estimating any of this backup by directly generating future states.
% Since UHM directly samples from the $n$-step future state distribution for any given $n$, it enables utilizing any sub-probability measure $\nu$. 
Among many possible choices of $\nu$, we present a value learning method using the winsorized geometric measure, which is defined as follows:
\begin{equation}
    \nu(k) =
    \begin{cases}
        (1-\lambda)(\lambda\gamma)^{k-1}, &\text{if $1 \le k < k_{\max}$,}\\
        (\lambda\gamma)^{k_{\max}-1}, &\text{if $k = k_{\max}$,}\\
        0, &\text{if $k > k_{\max}$,}
    \end{cases} \label{eq_winsor_geom_measure}
\end{equation}
which ensures the convergence guarantee in Proposition \ref{proposition_uhm_convergence} since it is a sub-probability that satisfies
% $\sum_{k\ge1}\nu(k)=1-\frac{\lambda-\lambda\gamma}{1-\lambda\gamma}\left\{1-(\lambda\gamma)^{k_{\max}-1}\right\}\le1$
$\sum_{k\ge1}\nu(k)\le1$. The corresponding $\xi^\nu$~(\ref{eq_xi}) yields
\begin{equation}
    \xi^\nu(k) =
    \begin{cases}
        \lambda(\lambda\gamma)^{k-1}, &\text{if $1 \le k < k_{\max}$,}\\
        0, &\text{if $k \ge k_{\max}$.}
    \end{cases}    
\end{equation}
This choice retains the original form of TD($\lambda$) while clipping excessively large and rare horizons, which in turn stabilizes the learning process.
% This choice retains the benefit of TD($\lambda$) as a flexible mixture of $n$-step TD targets, while clipping excessively large and rare horizons,
% leading to more stable learning.
For the horizon distribution $p_H$, we use the winsorized geometric distribution corresponding to the winsorized geometric measure~(\ref{eq_winsor_geom_measure}):
\begin{align}
    n' &\sim \text{Geom}(1-\lambda\gamma), n = \min(n', k_{\max}).
\end{align}
Then, we can obtain the value learning target $G^\nu$ (\ref{eq_uhm_td_target}) by substituting importance ratios:
\begin{align}
\label{eq_uhm_target_weights}
    w_\xi &=
    \begin{cases}
        \frac{\lambda}{1-\lambda\gamma}, &\text{if $1\le n < k_{\max}$,} \\
        0, &\text{if $n= k_{\max}$,}
    \end{cases} \\
    w_\nu &=
    \begin{cases}
        \frac{1-\lambda}{1-\lambda\gamma}, &\text{if $1\le n < k_{\max}$,} \\
        1, &\text{if $n= k_{\max}$.}
    \end{cases}
\end{align}
In the following section, we provide a practical offline RL algorithm using this value learning target.
% We note that resulting value target $G^\nu$ aligns with $\gamma$-MVE, except for the effect of horizon capping. However, this target is not attainable with GHM, since GHM does not reveal timesteps and is restricted to a fixed geometric horizon. On the other hand, UHM enables this construction by explicitly modeling the horizon and allowing flexible horizon distributions.
\subsection{Practical Implementations} \label{subsec:imp}
We now present a practical offline RL algorithm using UHM, whose pseudocode is provided in Algorithm \ref{algo:rl_with_uhm}. For notation, $\theta$ denotes the parameters of all neural networks and $\bar{\theta}$ represent their exponential moving average (EMA). At each update step, we perform the following update based on transition tuples $(s, a, r, s', a')$ sampled from the dataset.

\textbf{$\lambda$ scheduling.} Since the UHM is trained via bootstrapping, its output for large $n$ is inaccurate in the early training stage. To resolve this issue, we perform scheduling as $\lambda=\frac{r\lambda_f}{1 - (1-r)\lambda_f},$ where $\lambda_f$ is the final trace value and $r\in[0,1]$ is a training progress. It enforces the effective horizon of UHM to increase linearly from 1 to $1/(1-\lambda_f\gamma)$, making the bootstrapping of UHM more stable. Based on the scheduled $\lambda$, the maximum imaginary horizon $k_{\text{max}}$ is decided as $\text{qgeom}(1-\lambda\gamma, q)$, which is a q-quantile of the geometric distribution $\text{Geom}(1-\lambda\gamma)$. As a result, the horizon $n$ is sampled from the winsorized geometric distribution, which we denote as $\min(\text{Geom}(1-\lambda\gamma), k_{\text{max}})$.

\textbf{Model learning.} We train a vector field $v_\theta$ to utilize flow-matching \cite{flow_matching_lipman} as a generative model for UHM, and follow coupled-CFM \cite{tdflow_farebrother} to construct the learning objective (\ref{eq_uhm_bootstrap}). If $n > 1$, we first sample noise $s_e^0\sim\gN(0,I)$ and next action $\tilde{a}'$ to predict an $n-1$ step future state $s_e^1$, which is obtained by solving discretized ODE $s_e^{\tau + \Delta\tau} = x^\tau + v_{\bar{\theta}}(s_e^\tau|s',\tilde{a}',n-1,\tau)\Delta\tau$ from $\tau=0$ to $\tau=1$. Then $s_e^0$ is reused for constructing conditional optimal transport paths $s_e^\tau = (1-\tau)s_e^0 + \tau s_e^1$ for flow timestep $\tau \sim \text{Unif}[0,1]$, resulting in the flow-matching loss $L^v=||v_\theta(s_e^\tau|s,a,n,\tau) - (s_e^1 - s_e^0)||^2_2$. If $n=1$, we skip solving the ODE and set $s_e^1=s'$. We note that EMA weights $\bar{\theta}$ are used for generating bootstrapping targets to stabilize the training. 
% Additionally, we train the model to produce the change in states following \citet{delta_iris_micheli}, which yields better performance in locomotion tasks.

\textbf{Behavior mixing.} While UHM alleviates error accumulation through direct future prediction, it can still be inaccurate when queried with unseen state–action pairs. To address this issue, we introduce a simple behavior mixing strategy that uses a dataset action $a'$ with probability $\beta$ when generating the bootstrapping target $s_e$. Specifically, we deploy stochastically mixed policy $\pi^\text{mix}=(1-\beta)\pi_\theta + \beta \delta(a')$ to sample the next actions $\tilde{a}'$. It limits the total variation divergence from the behavior policy, analogous to classical RL papers \cite{kakade2002approximately, ross2010efficient}. We observe that the optimal choice of $\beta$ varies across tasks; however, to avoid exhaustive hyperparameter tuning, we fix $\beta = 0.3$ in all main experiments. The detailed analysis on the effect of $\beta$ is provided in Section \ref{subsec:ablations}.

\AtEndEnvironment{algorithmic}{\vspace{-1pt}}
\begin{algorithm}[t]
  \caption{Offline RL with UHM}
  \label{algo:rl_with_uhm}
  \label{alg:uhm}
  \begin{algorithmic}
    \STATE {\bfseries Input:} UHM vector field $v_\theta$, actor $\mu_\theta$, critic $Q_\theta$, reward model $R_\theta$, offline dataset $\gD$, discount factor $\gamma$, actor noise scale $\sigma$, actor BC coefficient $\alpha$, behavior mixing coefficient $\beta$, EMA decay $\eta$
    \STATE $\bar{\theta} \leftarrow \theta$
    \FOR{$i=1$ {\bfseries to} $N_{\text{update}}$}
    \STATE $(s,a,r,s', a') \sim \gD$
    \STATE Schedule $\lambda$ and $k_{\text{max}}$
    \STATE \textcolor{gray}{//\,\,Sample future states for bootstrapping}
    \STATE $n \sim \min(\text{Geom}(1-\lambda\gamma), k_{\text{max}})$
    \STATE $\tilde{a}' \sim (1-\beta)\gN(\mu_{\bar{\theta}}(s'), \sigma^2I) + \beta \delta(a')$
    \STATE $s_e^0 \sim \gN(0,I), \, s_e^1 \leftarrow s_e^0$
    \FOR{$j=1$ {\bfseries to} $N_{\text{flow}}$}
    \STATE $s_e^1 \leftarrow s_e^1 + \frac{1}{N_{\text{flow}}}v
    _{\bar{\theta}}(s_e^1|s',\tilde{a}',n-1,\frac{j-1}{N_{\text{flow}}})$
    \ENDFOR
    \STATE \textbf{if} $n = 1$ \textbf{then} $s_e^1 \leftarrow s'$
    \STATE \textcolor{gray}{//\,\, UHM loss}
    \STATE $\tau \sim \text{Unif}[0,1], \, s_e^\tau \leftarrow (1-\tau)s_e^0 + \tau s_e^1$
    \STATE $L^v \leftarrow ||v_\theta(s_e^\tau|s,a,n,\tau) - (s_e^1-s_e^0)||^2$
    \STATE $a_e \sim \gN(\mu_{\bar{\theta}}(s_e^1), \sigma^2I)$
    \STATE \textcolor{gray}{//\,\, Actor-critic loss}
    \STATE Compute $w_\xi, w_\nu$ according to (\ref{eq_uhm_target_weights})
    \STATE $G^\nu \leftarrow r + \gamma(w_\xi R_{\text{sg}(\theta)}(s_e^1, a_e) + w_\nu Q_{\bar{\theta}}(s_e^1,a_e))$
    \STATE $L^Q \leftarrow (Q_\theta(s,a) - G^\nu)^2$ 
    \STATE $L^\pi \leftarrow \alpha||\mu_\theta(s)-a||^2_2 -Q_{\text{sg}(\theta)}(s,\mu_\theta(s))$
    \STATE \textcolor{gray}{//\,\, Reward loss}
    \STATE $L^R \leftarrow (R_\theta(s, a) - r)^2$
    \STATE \textcolor{gray}{//\,\, Update network parameters}
    \STATE $L \leftarrow L^v + L^Q + L^R + L^\pi$
    \STATE $\theta \leftarrow \theta - \nabla_\theta L, \,\,\bar{\theta} \leftarrow (1 - \eta)\bar{\theta} + \eta\theta$
    \ENDFOR
  \end{algorithmic}
\end{algorithm}

\textbf{Rewards and terminations.} For reward modeling, we train a neural network $r_\theta$ by minimizing the mean-squared error $L^R = (R_\theta(s,a) - r)^2$. To properly handle terminal states, we employ an augmented state representation that concatenates a terminal indicator with the state, and train UHM to sample from this augmented space. We further treat terminal states as absorbing states that only transition to themselves and yield zero reward. It stabilizes value learning by preventing UHM from generating unseen combinations of states and terminal indicators. For notational simplicity, we do not introduce a separate symbol for the augmented state in Algorithm \ref{algo:rl_with_uhm}; however, explicitly modeling terminations and preventing value bootstrapping at terminal states are crucial for performance, as demonstrated in Section \ref{subsec:ablations}.

\textbf{Actor-critic learning.} We train a critic network $Q_\theta$ to minimize the TD learning objective $L^Q=(Q_\theta(s,a) - G^\nu)^2$, where $G^\nu$ is computed according to Equation~(\ref{eq_uhm_td_target}) using the target network $Q_{\bar{\theta}}$. For actor learning, we train a deterministic actor network $\mu_\theta$ to minimize the TD3+BC \cite{td3bc_fujimoto} objective $L^\pi=\alpha \lVert \mu_\theta(s) - a \rVert_2^2 - Q_{\text{sg}(\theta)}(s, \mu_\theta(s))$, where $\text{sg}(\cdot)$ denotes the stop-gradient operator. During training, we apply target smoothing by adding Gaussian noise with standard deviation $\sigma$, resulting in a stochastic policy $\pi_\theta(\cdot \mid s) = \mathcal{N}(\mu_\theta(s), \sigma^2 I)$.

\section{Experiments}
We conduct extensive experiments to evaluate the proposed offline RL algorithm. In Section \ref{subsec:exp_offline_rl}, we benchmark our method against competitive offline RL baselines across a diverse set of reward-based tasks in OGBench \cite{ogbench_park2025}, including tasks that provide highly suboptimal datasets or require long-horizon reasoning. In Section \ref{subsec:ablations}, we conduct a series of analyses to examine how individual components of the proposed method contribute to overall performance.

\subsection{Experiments on Offline RL Benchmarks}
\label{subsec:exp_offline_rl}
To evaluate the performance of the proposed method, we first conduct experiments on standard OGBench tasks that are commonly used in prior work~\cite{fql_park, chunking_rl_li, chen2025unleashing}. We then study more challenging settings to examine the limits of our approach, including (1) noisy tasks with highly suboptimal datasets and (2) long-horizon reasoning tasks that require substantially more interaction steps to solve. For readability, we omit the ``-singletask'' suffix and shorten task names by using task indices (e.g., \texttt{cube-single-play-singletask-task1-v0} is denoted as \texttt{cube-single-play-1}).

\subsubsection{Experimental Setup}
\textbf{Baselines.} We compare the proposed method with several baselines. For model-free methods, we use \textit{IQL} \cite{iql_kostrikov} that trains a critic with in-sample maximization, \textit{ReBRAC} \cite{rebrac_tarasov} that performs behavior-regularization for actor and critic updates, and \textit{FQL} \cite{fql_park} that trains a one-step flow policy regularized with behavior flow-matching policy. For model-based methods, we use \textit{MOPO} \cite{mopo_yu} that penalizes  rewards in predicted states with dynamics uncertainty, \textit{MOBILE} \cite{mobile_sun} that replaces dynamics uncertainty in MOPO with a disagreement in critic targets, and \textit{MAC} \cite{mac_park} that utilizes action chunking to prevent error accumulation and apply TD-$n$ with synthetic rollouts.

\textbf{Additional baselines.} We also compare our method with additional baselines, which are designed for rigorous ablation studies. We first tune actor BC coefficient $\alpha$ for ReBRAC again and denote it \textit{ReBRAC$^\dagger$}. Since it performs the same policy extraction method as our methods, we can examine whether our critic learning with UHM is effective by comparing with it. To assess the advantage of UHM over single-step dynamics models, we use \textit{MBTD($\lambda$)}, which generates synthetic rollouts using a single-step flow dynamics model to perform TD($\lambda$) analogous to LEQ \cite{leq_park}. We also compare against \textit{DTD($\lambda$)}, which performs TD($\lambda$) over trajectories sampled directly from the dataset, to determine whether on-policy value learning is necessary for performance. Finally, we compare our method with \textit{GHM} that trains a critic with $\gamma$-MVE, to verify whether the increased flexibility of UHM provides tangible benefits in the offline RL setting. For fair comparison, all additional baselines use the same hyperparameters and design choices as the proposed method. Only actor BC coefficient $\alpha$ is tuned separately for each method.

\textbf{Evaluation.}
For the standard OGBench tasks, we report the performance of model-free baselines from \citet{fql_park} and model-based baselines from \citet{mac_park}. We additionally tune the model-based baselines for locomotion tasks following the same experimental protocol, as it is omitted in the paper. Each experiment is run with five random seeds, and we report the average success rate along with the standard deviation. All agents share the same network architectures and are trained for a total of 1M gradient steps. Performance is evaluated by averaging the results over the last three evaluation epochs. For both the proposed method and the additional baselines, the final value of the trace parameter $\lambda_f$ is set to 0.8 and the discount factor is set to 0.999.

For noisy and long-horizon reasoning tasks, we report results only for the proposed method and the additional baselines, as these methods consistently outperform other approaches on the standard tasks. For noisy tasks, we follow the same evaluation protocol as in the standard setting. For long-horizon tasks, we use three random seeds and train agents for 2M gradient steps, while increasing $\lambda_f$ to 0.9 to encourage horizon reduction. The dataset for the long-horizon reasoning tasks consists of 10M transitions, which is ten times larger than that used for the standard and noisy tasks. Please refer to Appendix \ref{sec:appendix_exp_details} for more detailed experimental setup.

\begin{table*}[ht]
\caption{Results on 50 standard tasks in OGBench. We omit the standard deviations of average success rates for methods whose results are taken from prior works \cite{fql_park, mac_park}. Please refer to Table \ref{ogbench_regular_full_table} for the detailed per-task results.}
\label{ogbench_regular_table}
\centering
\small
% Resize the table to 1\textwidth (the full width of the text area)
\resizebox{\textwidth}{!}{
    \begin{tabular}{lccccccccccc}
    \toprule
     & \multicolumn{3}{c}{\textbf{Model-Free}} 
     & \multicolumn{3}{c}{\textbf{Model-Based}}
     & \multicolumn{4}{c}{\textbf{Ablations}}
     & \multicolumn{1}{c}{\textbf{Ours}} \\
    \cmidrule(lr){2-4} \cmidrule(lr){5-7} \cmidrule(lr){8-11}\cmidrule{12-12}
     Environments (5 tasks each) & IQL & ReBRAC & FQL & MOPO & MOBILE & MAC & ReBRA$\text{C}^\dagger$ & MBTD($\lambda$) & DTD($\lambda$) & GHM & UHM \\
    \midrule
    antmaze-large-navigate  & 53 $\pm$ 3 & 81 $\pm$ 5 & 79 $\pm$ 3 & 0 $\pm$ 0 & 0 $\pm$ 0 & 18 $\pm$ 4 & 72 $\pm$ 14 & 71 $\pm$ 11 & \textbf{93} $\pm$ 1 & \underline{90} $\pm$ 2 & \underline{89} $\pm$ 1 \\
    antmaze-giant-navigate  & 4 $\pm$ 1 & 26 $\pm$ 8 & 9 $\pm$ 5 & 0 $\pm$ 0 & 0 $\pm$ 0 & 0 $\pm$ 0 & 30 $\pm$ 13 & 27 $\pm$ 3 & \textbf{52} $\pm$ 11 & 33 $\pm$ 7 & 36 $\pm$ 4 \\
    humanoidmaze-medium-navigate  & 33 $\pm$ 2 & 22 $\pm$ 8 & 58 $\pm$ 5 & 0 $\pm$ 0 & 0 $\pm$ 0 & 2 $\pm$ 0 & 22 $\pm$ 10 & 64 $\pm$ 11 & 81 $\pm$ 2 & 90 $\pm$ 1 & \textbf{95} $\pm$ 1 \\
    humanoidmaze-large-navigate  & 2 $\pm$ 1 & 2 $\pm$ 1 & 4 $\pm$ 2 & 0 $\pm$ 0 & 0 $\pm$ 0 & 0 $\pm$ 0 & 1 $\pm$ 1 & 16 $\pm$ 3 & 27 $\pm$ 10 & 16 $\pm$ 1 & \textbf{33} $\pm$ 9 \\
    antsoccer-arena-navigate  & 8 $\pm$ 2 & 0 $\pm$ 0 & \textbf{60} $\pm$ 2 & 0 $\pm$ 0 & 0 $\pm$ 0 & 29 $\pm$ 4 & 1 $\pm$ 1 & 47 $\pm$ 3 & 0 $\pm$ 0 & 20 $\pm$ 4 & 26 $\pm$ 4 \\
    cube-single-play  & 83 $\pm$ 3 & 91 $\pm$ 2 & \underline{96} $\pm$ 1 & 12 $\pm$ 4 & 81 $\pm$ 8 & \textbf{99} $\pm$ 2 & 91 $\pm$ 2 & 92 $\pm$ 1 & 90 $\pm$ 2 & 91 $\pm$ 3 & 92 $\pm$ 3 \\
    cube-double-play  & 7 $\pm$ 1 & 12 $\pm$ 1 & 29 $\pm$ 2 & 1 $\pm$ 1 & 1 $\pm$ 2 & \textbf{53} $\pm$ 4 & 4 $\pm$ 2 & 4 $\pm$ 1 & 4 $\pm$ 1 & 29 $\pm$ 1 & 30 $\pm$ 2 \\
    scene-play  & 28 $\pm$ 1 & 41 $\pm$ 3 & 56 $\pm$ 2 & 6 $\pm$ 8 & 8 $\pm$ 4 & \textbf{97} $\pm$ 4 & 40 $\pm$ 3 & 31 $\pm$ 2 & 76 $\pm$ 3 & 44 $\pm$ 3 & 43 $\pm$ 4 \\
    puzzle-3x3-play  & 9 $\pm$ 1 & 21 $\pm$ 1 & 30 $\pm$ 1 & 20 $\pm$ 0 & 12 $\pm$ 9 & 20 $\pm$ 0 & 90 $\pm$ 4 & 93 $\pm$ 2 & \textbf{99} $\pm$ 0 & 51 $\pm$ 2 & \textbf{99} $\pm$ 1 \\
    puzzle-4x4-play  & 7 $\pm$ 1 & 14 $\pm$ 1 & 17 $\pm$ 2 & 0 $\pm$ 0 & 0 $\pm$ 0 & \textbf{78} $\pm$ 13 & 1 $\pm$ 0 & 4 $\pm$ 0 & 1 $\pm$ 0 & 13 $\pm$ 1 & 11 $\pm$ 2 \\
    \midrule
    Average  & 23 & 31 & 44 & 4 & 10 & 40 & 35 $\pm$ 2 & 45 $\pm$ 2 & 52 $\pm$ 1 & 48 $\pm$ 1 & \textbf{55} $\pm$ 1 \\
    \bottomrule
    \end{tabular}
}
\end{table*}
\begin{table}[t]
\caption{Results on 25 noisy tasks in OGBench. Please refer to Table \ref{ogbench_stitching_full_table} for the detailed per-task results.}
\label{ogbench_stitching_table}
\centering
\small
% Resize the table to 1\textwidth (the full width of the text area)
\resizebox{\linewidth}{!}{
    \begin{tabular}{lccccc}
    \toprule
     Environments (5 tasks each) & ReBRA$\text{C}^\dagger$ & MBTD($\lambda$) & DTD($\lambda$) & GHM & UHM \\
    \midrule
    antmaze-medium-explore  & 74 $\pm$ 5 & \textbf{96} $\pm$ 2 & 81 $\pm$ 6 & 91 $\pm$ 3 & 89 $\pm$ 4 \\
    antmaze-large-explore  & 13 $\pm$ 7 & 20 $\pm$ 2 & 18 $\pm$ 9 & 14 $\pm$ 2 & \textbf{26} $\pm$ 5 \\
    cube-double-noisy  & 2 $\pm$ 1 & 2 $\pm$ 1 & 2 $\pm$ 2 & \textbf{24} $\pm$ 3 & 18 $\pm$ 1 \\
    scene-noisy  & 27 $\pm$ 7 & 42 $\pm$ 2 & 6 $\pm$ 3 & 57 $\pm$ 3 & \textbf{61} $\pm$ 3 \\
    puzzle-4x4-noisy  & 0 $\pm$ 0 & 0 $\pm$ 0 & 0 $\pm$ 0 & \textbf{4} $\pm$ 1 & 1 $\pm$ 0 \\
    \midrule
    Average  & 23 $\pm$ 2 & 32 $\pm$ 1 & 23 $\pm$ 4 & \underline{38} $\pm$ 1 & \textbf{39} $\pm$ 1 \\
    \bottomrule
    \end{tabular}
}
\end{table}

\subsubsection{Results}
We now report experimental results across the three task categories. In all tables, the best-performing baseline is highlighted in bold. We additionally underline results that achieve at least 95\% of the best success rate.

\textbf{Results on standard tasks.}
Table \ref{ogbench_regular_table} reports the mean success rates and standard deviations across the standard OGBench tasks. Overall, model-based baselines exhibit weaker performance than model-free baselines across most tasks. In particular, MOPO and MOBILE achieve success rates below 10\% on nearly all tasks, suggesting that uncertainty-based reward penalties may significantly hinder value learning in sparse-reward settings. MAC significantly outperforms other baselines on several manipulation tasks, but performs poorly on locomotion tasks.
We hypothesize that this limitation arises from rejection sampling over the behavior policy, which may struggle to handle high-dimensional actions.

Among the additional baselines, methods that leverage $n$-step returns for TD learning outperform one-step TD methods, such as ReBRAC$^\dagger$ and FQL. This highlights the importance of horizon reduction in value learning. Notably, although MBTD($\lambda$) and GHM perform theoretically sound value learning via model-based value expansion, they achieve lower performance than DTD($\lambda$), which uses dataset trajectories. In contrast to these approaches, UHM is the only method that outperforms DTD($\lambda$). We attribute this advantage to directly predicting $n$-step future states while capping the maximum future horizon, which makes value learning more scalable and robust.

\begin{table}[t]
\caption{Results on 25 long-horizon reasoning tasks in OGBench. Please refer to Table \ref{ogbench_long_horizon_full_table} for the detailed per-task results.}
\label{ogbench_long_horizon_table}
\centering
\small
% Resize the table to 1\textwidth (the full width of the text area)
\resizebox{\linewidth}{!}{
    \begin{tabular}{lccccc}
    \toprule
     Environments (5 tasks each) & ReBRA$\text{C}^\dagger$ & MBTD($\lambda$) & DTD($\lambda$) & GHM & UHM \\
    \midrule
    cube-triple-play  & 4 $\pm$ 1 & 12 $\pm$ 2 & 1 $\pm$ 1 & 44 $\pm$ 5 & \textbf{56} $\pm$ 3 \\
    cube-quadruple-play  & 0 $\pm$ 0 & 0 $\pm$ 0 & 0 $\pm$ 1 & 12 $\pm$ 4 & \textbf{16} $\pm$ 6 \\
    puzzle-4x5-play  & 7 $\pm$ 2 & 5 $\pm$ 3 & 14 $\pm$ 3 & \textbf{17} $\pm$ 1 & \underline{16} $\pm$ 0 \\
    puzzle-4x6-play  & \textbf{11} $\pm$ 5 & 1 $\pm$ 1 & 5 $\pm$ 6 & 5 $\pm$ 3 & \textbf{11} $\pm$ 1 \\
    humanoidmaze-giant-navigate  & 2 $\pm$ 1 & 5 $\pm$ 1 & \textbf{46} $\pm$ 5 & 4 $\pm$ 0 & 10 $\pm$ 2 \\
    \midrule
    Average  & 5 $\pm$ 1 & 5 $\pm$ 1 & 13 $\pm$ 3 & 16 $\pm$ 2 & \textbf{22} $\pm$ 1 \\
    \bottomrule
    \end{tabular}
}
\end{table}

\textbf{Results on noisy tasks.} In Table~\ref{ogbench_stitching_table}, we report the mean and standard deviations of success rates across 25 noisy tasks in OGBench. Compared to standard tasks, DTD($\lambda$) struggles to learn effective policies and even exhibits comparable average success rates with ReBRAC$^\dagger$. It achieves success rates below 10\% on all manipulation environments, which is substantially lower than its performance on standard tasks. This observation suggests that relying on trajectories from highly suboptimal datasets can severely hinder effective value learning. On the other hand, model-based approaches consistently outperform model-free methods on noisy tasks. We attribute this improvement to the use of imagined on-policy trajectories rather than directly relying on suboptimal transitions from the dataset. Among model-based approaches, GHM and UHM outperform MBTD($\lambda$), with UHM achieving the best overall performance. These results demonstrate the scalability of the proposed method to offline RL tasks with highly suboptimal data.

\textbf{Results on long-horizon reasoning tasks.} In Table~\ref{ogbench_long_horizon_table}, we report performance on long-horizon reasoning tasks in OGBench. Both GHM and UHM achieve stronger performance than DTD($\lambda$), suggesting that relying on dataset trajectories for long-horizon returns can lead to increased distributional mismatch. MBTD($\lambda$) performs comparably to ReBRAC$^\dagger$, failing to yield meaningful gains from long-horizon imagination when using single-step dynamics models. Among the compared methods, UHM achieves the best overall performance, improving the average success rate by approximately 69\% over DTD($\lambda$) and 38\% over GHM, demonstrating its scalability to long-horizon reasoning tasks. However, in \texttt{humanoidmaze-giant}, DTD($\lambda$) significantly outperforms model-based methods, suggesting that accurate model learning remains challenging in large-scale, high-dimensional domains. This result highlights the need for future research that combines our approach with techniques for handling high-dimensional observations and for reducing the effective decision horizon of the policy.

To summarize, our experimental results reveal three key findings. 
First, horizon reduction plays a crucial role in effective value learning. 
Second, model-based value expansion without repeated inference scales well to noisy tasks and long-horizon reasoning tasks. 
Finally, the proposed method shows competitive performance across the task categories we consider, with a 14\% higher average success rate than the second-best method.

\subsection{Ablation Studies}
\label{subsec:ablations}
In this section, we provide ablation studies to analyze the contribution of each component of the proposed method. In all figures, shaded areas denote a standard deviation over five independent runs.

\begin{figure}[t]
  \centering
  \includegraphics[width=1.0\linewidth]{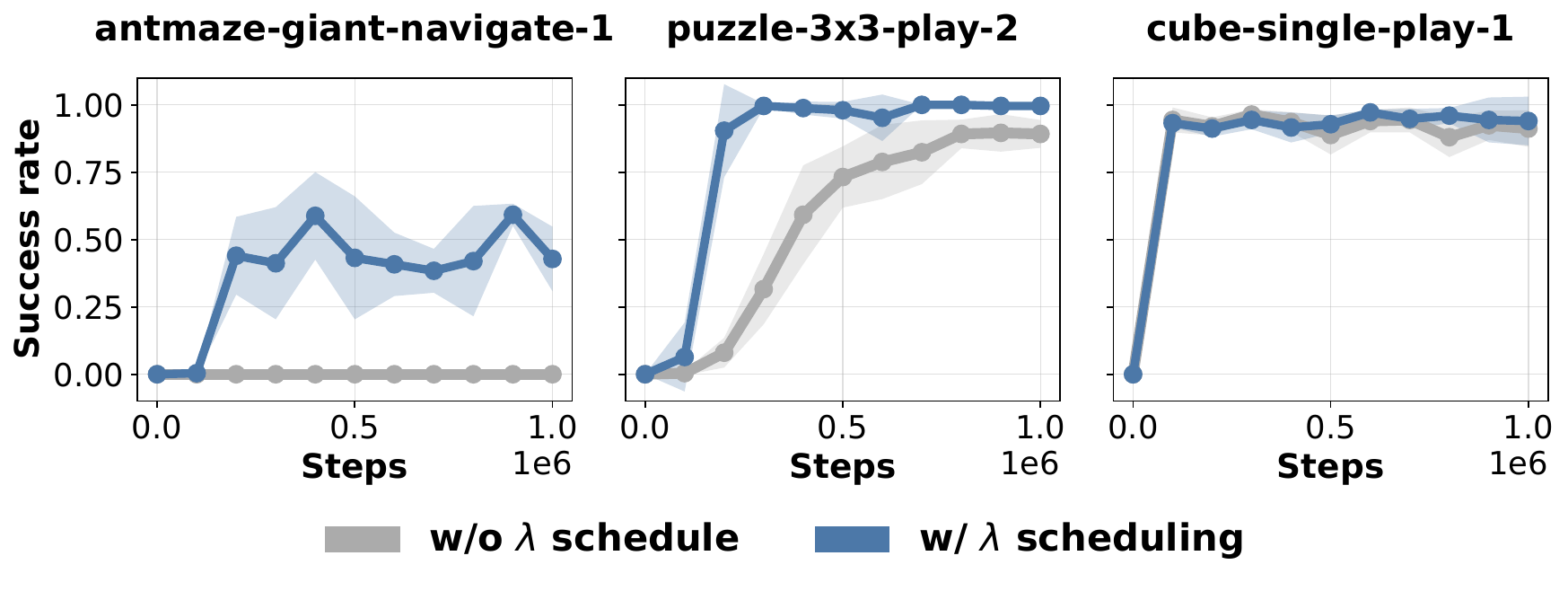} 
\caption{Learning curves with and without $\lambda$ scheduling.}
\label{fig:scheduling_ablation}
\end{figure}
\begin{figure}[t]
  \centering
  \includegraphics[width=1.0\linewidth]{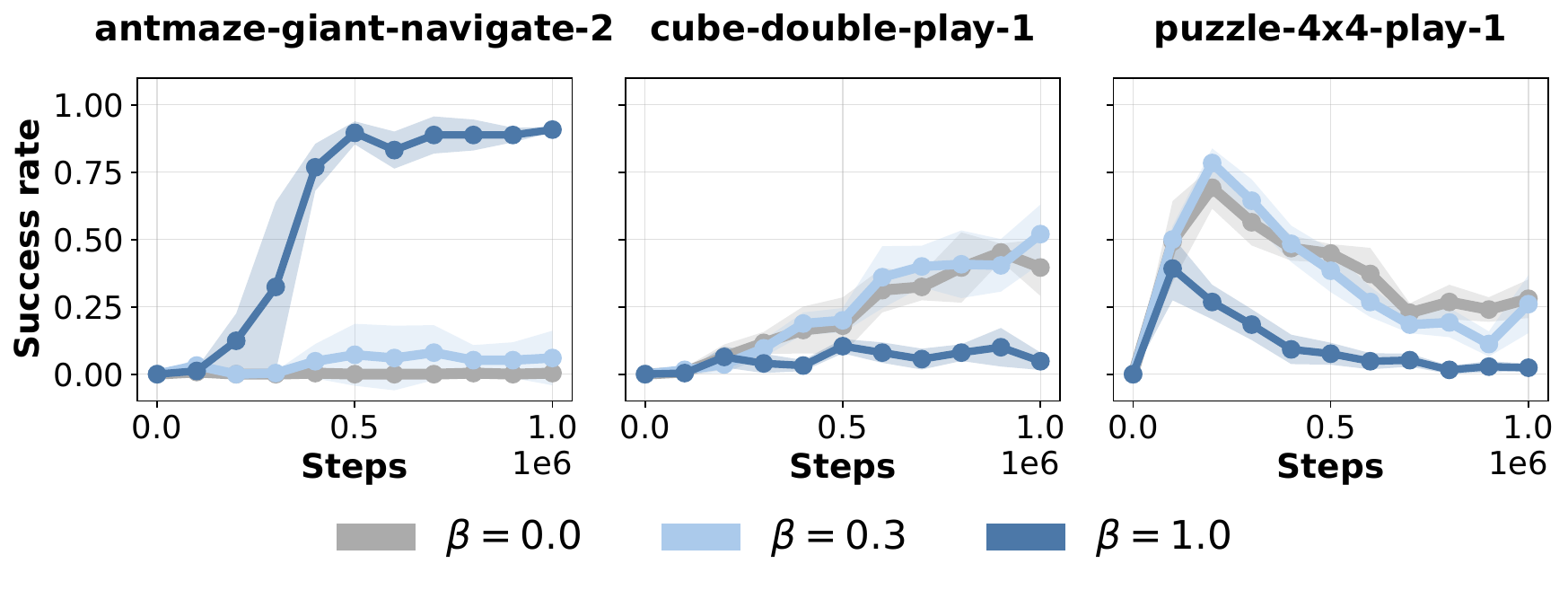} 
\caption{Learning curves for different behavior mixing coefficients $\beta$.}
\label{fig:bm_ablation}
\end{figure}

\textbf{Ablations on $\lambda$ scheduling.} Figure~\ref{fig:scheduling_ablation} compares the learning curves of the proposed method with and without $\lambda$ scheduling. Empirically, $\lambda$ scheduling consistently improves performance across nearly all tasks. In \texttt{antmaze-giant-navigate-1}, the method fails to learn meaningful policies without scheduling, whereas introducing $\lambda$ scheduling enables better performance. Even in relatively easier tasks, $\lambda$ scheduling yields consistent performance gains and faster convergence. These results suggest that $\lambda$ scheduling enables efficient value learning.

\begin{figure}[t]
  \centering
  \includegraphics[width=1.0\linewidth]{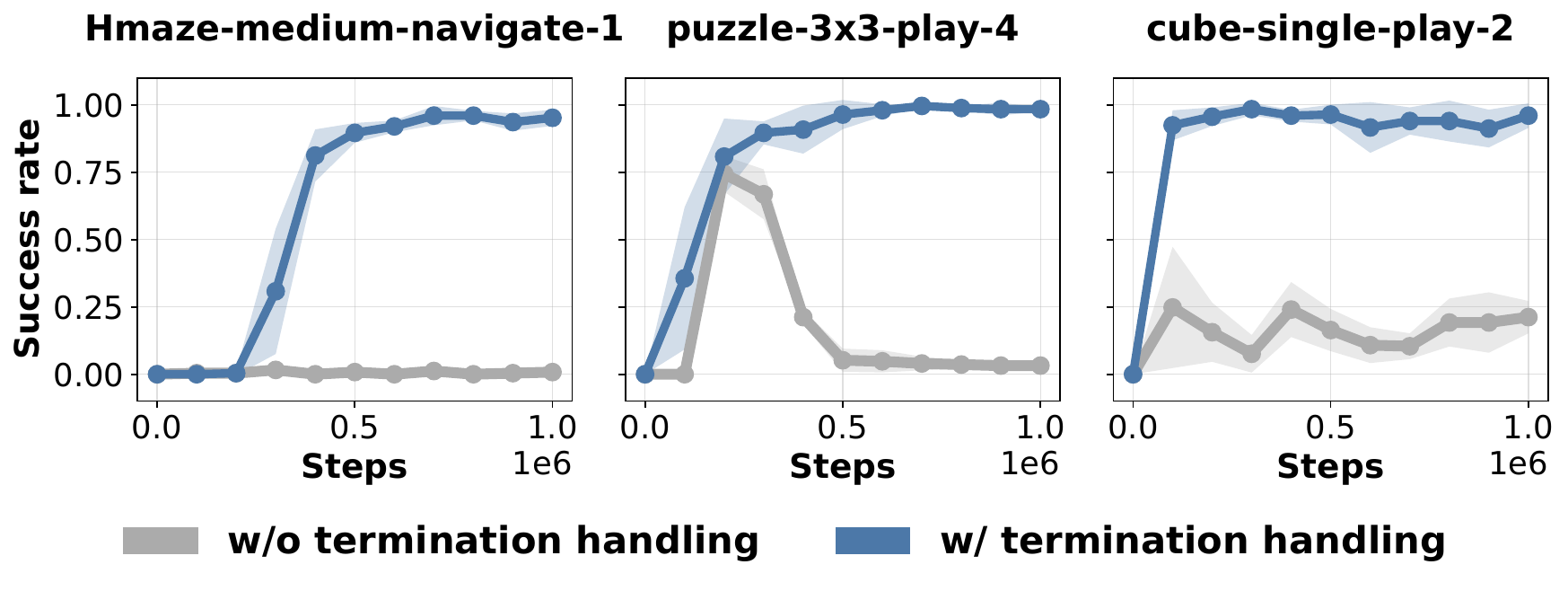} 
\caption{Learning curves with and without terminal state handling.}
\label{fig:termination}
\end{figure}
\begin{figure}[t]
  \centering
  \includegraphics[width=1.0\linewidth]{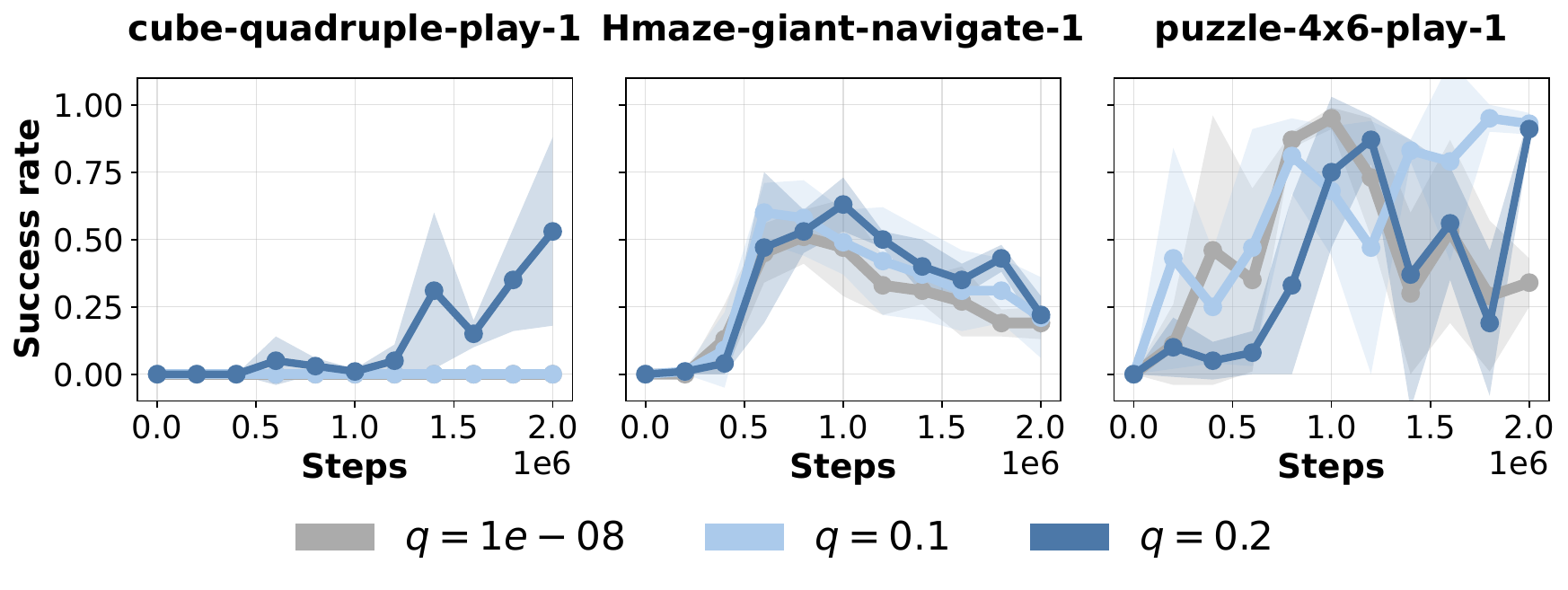} 
\caption{Learning curves for different horizon winsorization quantiles $q$.}
\label{fig:q_ablation}
\end{figure}

\begin{figure}[t]
  \centering
  \includegraphics[width=1.0\linewidth]{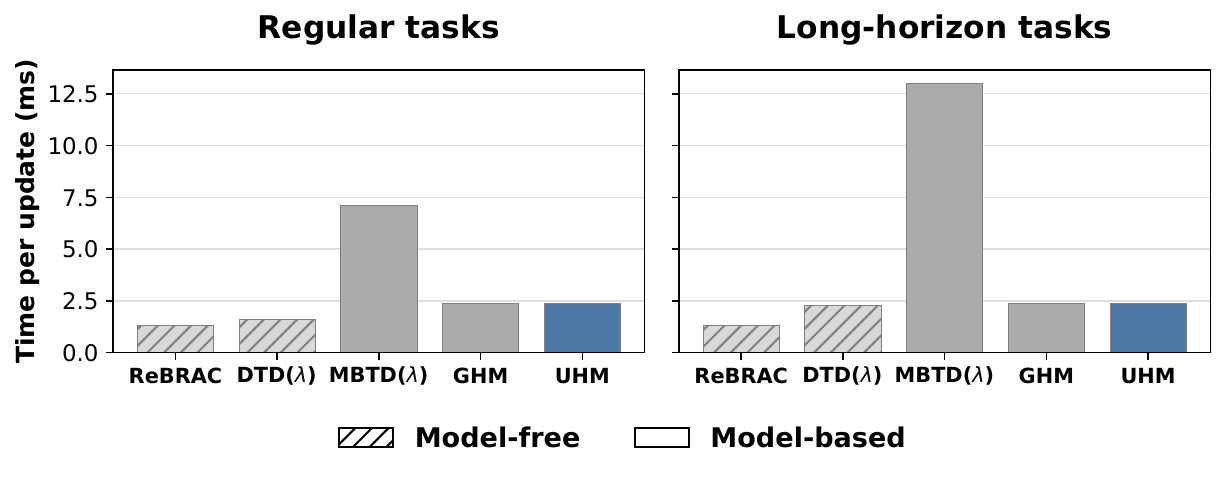} 
\caption{Wall-clock time per gradient update across baselines.}
\label{fig:update_time}
\end{figure}

\textbf{Ablations on the behavior mixing coefficient $\beta$.} Figure \ref{fig:bm_ablation} illustrates the effect of the behavior mixing coefficient $\beta$. We observe that the optimal value of $\beta$ varies across tasks. In \texttt{antmaze-giant-navigate-2}, setting $\beta=1.0$ is crucial for stable learning, whereas smaller values of $\beta$ lead to poor performance. In contrast, tasks such as \texttt{cube-double-play-1} and \texttt{puzzle-4x4-play-1} benefit from smaller values of $\beta$. Meanwhile, performance averaged over all representative tasks from the standard environments remains relatively robust to the choice of $\beta$, with success rates of 0.63, 0.66, and 0.59 for $\beta=0.0, 0.3$, and $1.0$, respectively. Based on these observations, we recommend tuning $\beta$ to achieve optimal task-specific performance, while we fix $\beta=0.3$ in our main experiments for simplicity.

\textbf{Ablations on terminal state handling.} Figure~\ref{fig:termination} illustrates the effect of explicitly handling terminal states. Ignoring terminal states leads to a substantial degradation in performance across all evaluated tasks. We can infer that bootstrapping value estimates from terminal states causes unstable learning. These results highlight that handling terminal states is crucial for the performance.

\textbf{Ablations on the horizon winsorization quantile $q$.} Figure \ref{fig:q_ablation} illustrates the effect of the winsorization quantile $q$ on a subset of long-horizon reasoning tasks. Overall, $q=0.1$ and $q=0.2$ outperform $q=10^{-8}$, indicating that winsorization is important for performance. However, the best choice of $q$ can be task-dependent: \texttt{cube-quadruple-play-1} performs well only with $q=0.2$, while \texttt{puzzle-4x6-play-1} is much more stable with $q=0.1$. We therefore fix $q=0.2$ in the main experiments for simplicity, although tuning $q$ may further improve performance.

\textbf{Update time comparison.} Figure~\ref{fig:update_time} compares the wall-clock time required for a single gradient update with an RTX 4090 GPU. While model-free approaches demonstrate the lowest update time, MBTD($\lambda$) exhibits substantially higher update time, as it relies on repeated inference of a dynamics model. Notably, GHM and UHM significantly reduce this overhead by directly predicting future states. This advantage is more pronounced in long-horizon tasks, where UHM achieves update times within 10\% of DTD($\lambda$) while retaining the benefits of model-based value expansion.

\section{Conclusion}
In this work, we introduce universal horizon models (UHM), which generalize geometric horizon models by allowing the future horizon to be sampled from arbitrary distributions. We further propose a scalable algorithm for offline model-based value expansion and demonstrate that our method outperforms baselines across diverse tasks in OGBench. Despite these results, our approach has several limitations. First, the accuracy of UHM is constrained by data scarcity, which can lead the model to extrapolate beyond the data support. It suggests the need for additional mechanisms to guide predictions toward in-distribution states. Second, the performance of UHM is limited by model capacity, as accurately modeling longer horizons may require more expressive network architectures. Addressing these limitations and extending UHM to handle visual observations with action chunks remain interesting directions for future work.

\section*{Acknowledgments}
% This work was supported by Institute of Information \& Communications Technology Planning \& Evaluation (IITP) grant funded by the Korea government (MSIT) (No. 2019-0-01190, [SW Star Lab] Robot Learning: Efficient, Safe, and Socially-Acceptable Machine Learning).
This work was partly supported by Institute of Information \& Communications Technology Planning
\& Evaluation (IITP) grant funded by the Korea government (MSIT) (No. 2019-0-01190, [SW Star
Lab] Robot Learning: Efficient, Safe, and Socially-Acceptable Machine Learning, 50\%), and Institute of Information \& communications Technology Planning \& Evaluation (IITP) grant funded by the Korea government(MSIT)
(NO.RS-2021-II211343, Artificial Intelligence Graduate School Program [Seoul National University],
50\%).

\section*{Impact Statement}
This paper presents work whose goal is to advance the field of Machine
Learning. There are many potential societal consequences of our work, none of which we feel must be specifically highlighted here.

% \clearpage
\raggedbottom


\bibliography{reference}
\bibliographystyle{icml2026}

%%%%%%%%%%%%%%%%%%%%%%%%%%%%%%%%%%%%%%%%%%%%%%%%%%%%%%%%%%%%%%%%%%%%%%%%%%%%%%%
%%%%%%%%%%%%%%%%%%%%%%%%%%%%%%%%%%%%%%%%%%%%%%%%%%%%%%%%%%%%%%%%%%%%%%%%%%%%%%%
% APPENDIX
%%%%%%%%%%%%%%%%%%%%%%%%%%%%%%%%%%%%%%%%%%%%%%%%%%%%%%%%%%%%%%%%%%%%%%%%%%%%%%%
%%%%%%%%%%%%%%%%%%%%%%%%%%%%%%%%%%%%%%%%%%%%%%%%%%%%%%%%%%%%%%%%%%%%%%%%%%%%%%%
\newpage
\appendix
\onecolumn
\section{Theoretical Analysis}
\subsection{Derivation of Equation \ref{eq_gamma_mve}}
\label{subsec:td_lambda and gamma_mve}
$\gamma$-MVE \cite{gamma_janner} is originally proposed to conduct value learning after a $H$-step GHM rollout:
\begin{equation}
\label{eq_original_gamma_mve}
Q_{\gamma\text{-MVE}} = R(s,a) + \frac{\gamma}{1-\gamma}\sum_{n=1}^H\frac{(1-\gamma)(\gamma-\tilde{\gamma})^{n-1}}{(1-\tilde{\gamma})^n}\mathbb{E}_{\substack{s_e \sim m^\pi_n(\cdot\vert s,a) \\ a_e \sim \pi(\cdot|s_e)}} \left[ R(s_e,a_e) \right] + \gamma\bigg(\frac{\gamma - \tilde{\gamma}}{1-\tilde{\gamma}}\bigg)^H \mathbb{E}_{\substack{s_e \sim m^\pi_H(\cdot\vert s,a) \\ a_e \sim \pi(\cdot|s_e)}}\left[Q(s_e,a_e)\right],
\end{equation}
where $m^\pi_n$ denotes a distribution over states at the $n$th sequential step of a GHM rollout, and $\tilde{\gamma} \in (0,\gamma]$ denotes a discount factor to train the GHM. For the single-step rollout case, the equation (\ref{eq_original_gamma_mve}) is simplified as follows:
\begin{equation}
\label{eq_gamma_mve_appendix}
Q_{\gamma\text{-MVE}} = R(s,a) + \gamma\mathbb{E}_{\substack{s_e \sim m^\pi(\cdot\vert s,a) \\ a_e \sim \pi(\cdot|s_e)}} \bigg[ \frac{1}{1-\tilde{\gamma}}R(s_e, a_e) + \frac{\gamma - \tilde{\gamma}}{1-\tilde{\gamma}}Q(s_e,a_e) \bigg].
\end{equation}
By defining $\lambda:=\tilde{\gamma}/\gamma$, the equation (\ref{eq_gamma_mve_appendix}) can be rewritten as follows:
\begin{equation}
\label{eq_gamma_mve_lambda_appendix}
Q_{\gamma\text{-MVE}} = R(s,a) + \gamma\mathbb{E}_{\substack{s_e \sim m^\pi(\cdot\vert s,a) \\ a_e \sim \pi(\cdot|s_e)}} \bigg[ \frac{1}{1-\lambda\gamma}R(s_e, a_e) + \frac{(1-\lambda)\gamma}{1-\lambda\gamma}Q(s_e,a_e) \bigg].
\end{equation}

Using the definition of GHM $m^\pi(s_e|s,a) = (1-\lambda\gamma)\sum_{k=0}^\infty (\lambda\gamma)^k\text{Pr}(s_{k+1}=s_e|s_0=s, a_0=a, \pi)$, we can derive that expectations of $\gamma$-MVE target and TD($\lambda$) target are equal.
\begin{align}
    Q_{\gamma\text{-MVE}} &= \sum_{k=0}^{\infty}\{R(s_e,a_e) + (1-\lambda)\gamma Q(s_e,a_e)\}(\lambda\gamma)^k\text{Pr}(s_{k+1}=s_e, a_k=a_e| s_0=s,a_0=a,\pi) \\
    &= \mathbb{E}\bigg[\sum_{k=0}^\infty (\lambda\gamma)^kR_{k+1} + (1-\lambda)\gamma\sum_{k=0}^\infty (\lambda\gamma)^k Q(s_{k+1}, a_{k+1}) \,\,\bigg\vert\,\, s_0=s,a_0=a, \pi\bigg] \\
    &= \mathbb{E}\bigg[(1-\lambda)\sum_{k=0}^\infty \lambda^k(\sum_{l=0}^k \gamma^l R_{l+1} + \gamma^{k+1}Q(s_{k+1}, a_{k+1}))  \,\,\bigg\vert\,\, s_0=s,a_0=a, \pi\bigg] \\
    &= \mathbb{E}\bigg[(1-\lambda)\sum_{k=0}^\infty \lambda^kG_t^{(k+1)}  \,\,\bigg\vert\,\, s_0=s,a_0=a, \pi\bigg] \\
    &= \mathbb{E}\bigg[G_t^\lambda  \,\,\bigg\vert\,\, s_0=s,a_0=a, \pi\bigg].
\end{align}
This result indicates that GHM realizes TD($\lambda$) with single-step rollouts, avoiding the need for recursive inference inherent to methods based on single-step dynamics models.

\subsection{Proof of Proposition \ref{proposition_uhm_convergence}}
\label{subsec:proposition proof}
% \begin{proof}
Define the Banach space $\gB(\gS\times\gA)$ as the set of all bounded functions $Q:\gS\times\gA\to\mathbb{R}$ equipped with the supremum norm $\|Q\|_\infty := \sup_{(s,a)\in\gS\times\gA}|Q(s,a)|$.
Let $\nu$ be a sub-probability measure on $\mathbb{N}$, i.e., $\nu(k)\ge 0$ and $\sum_{k\ge 1}\nu(k)\le 1$.
Recall the $\nu$-Bellman operator $\gT^\nu$ defined by
\begin{align}
(\gT^\nu Q)(s,a)
:= \E\Big[
R(s,a)
+ \gamma \sum_{k\ge 1}\big(\xi^\nu(k)R(s_{k},a_{k}) + \nu(k)Q(s_{k},a_{k})\big)
\;\Big|\; s_0=s, a_0=a, \pi
\Big].
\end{align}
% where
% \begin{align}
% \xi^\nu(k) := \gamma^{k-1} - \sum_{\kappa=0}^{k-1}\gamma^\kappa \nu(k-\kappa).
% \end{align}
For any $Q,Q'\in \gB(\gS\times\gA)$ and any $(s,a)\in\gS\times\gA$,
\begin{align}
\big|(\gT^\nu Q)(s,a) - (\gT^\nu Q')(s,a)\big|
&= \gamma \Big|\E\Big[\sum_{k\ge 1}\nu(k)\big(Q(s_{k},a_{k})-Q'(s_{k},a_{k})\big)\,\Big|\, s_0=s,a_0=a,\pi\Big]\Big| \\
&\le \gamma\, \E\Big[\sum_{k\ge 1}\nu(k)\,\big|Q(s_{k},a_{k})-Q'(s_{k},a_{k})\big| \,\Big|\, s_0=s,a_0=a,\pi\Big] \\
&\le \gamma \Big(\sum_{k\ge 1}\nu(k)\Big)\|Q-Q'\|_\infty \\
&\le \gamma \|Q-Q'\|_\infty.
\end{align}
Taking the supremum over $(s,a)$ yields
\begin{align}
\|\gT^\nu Q - \gT^\nu Q'\|_\infty \le \gamma \|Q-Q'\|_\infty,
\end{align}
so $\gT^\nu$ is a $\gamma$-contraction. The Banach fixed point theorem implies that $\gT^\nu$ admits a unique fixed point $Q^\star\in \gB(\gS\times\gA)$, and the iteration $Q_{n+1}=\gT^\nu Q_n$ converges in $\|\cdot\|_\infty$ to $Q^\star$ for any initial $Q_0\in \gB(\gS\times\gA)$.

Recall
\begin{align}
Q^\pi(s,a) = \E\Big[\sum_{j\ge 0}\gamma^j R(s_{j},a_{j}) \,\Big|\, s_0=s,a_0=a,\pi\Big].
\end{align}
Substituting $Q^\pi$ into $\gT^\nu$ gives
\begin{align}
(\gT^\nu Q^\pi)(s,a)
&= \E\Big[
R(s,a)
+ \gamma \sum_{k\ge 1}\xi^\nu(k)R(s_{k},a_{k})
+ \gamma \sum_{i\ge 1}\nu(i)Q^\pi(s_{i},a_{i})
\;\Big|\; s_0=s, a_0=a, \pi
\Big] \\
&= \E\Big[
R(s,a)
+ \gamma \sum_{k\ge 1}\xi^\nu(k)R(s_{k},a_{k}) \nonumber\\
&\quad\quad\quad\quad\quad
+ \gamma \sum_{i\ge 1}\nu(i)\sum_{j\ge 0}\gamma^j R(s_{i+j},a_{i+j})
\;\Big|\; s_0=s, a_0=a, \pi
\Big].
\end{align}
Re-index the double sum by $k=i+j\ (\ge 1)$:
\begin{align}
\gamma \sum_{i\ge 1}\nu(i)\sum_{j\ge 0}\gamma^j R(s_{i+j},a_{i+j})
= \gamma\sum_{k\ge 1}\left(\sum_{i=1}^{k}\gamma^{k-i}\nu(i)\right) R(s_{k},a_{k}).
\end{align}
Therefore, for each $k\ge 1$, the coefficient of $R(s_{k},a_{k})$ equals
\begin{align}
\gamma \xi^\nu(k) + \gamma\sum_{i=1}^{k}\gamma^{k-i}\nu(i)=\gamma \gamma^{k-1}=\gamma^k.
\end{align}
Hence,
\begin{align}
(\gT^\nu Q^\pi)(s,a)
&= \E\Big[
R(s,a) + \sum_{k\ge 1}\gamma^k R(s_{k},a_{k})
\;\Big|\; s_0=s,a_0=a,\pi
\Big] \\
&= \E\Big[\sum_{k\ge 0}\gamma^k R(s_{k},a_{k}) \,\Big|\, s_0=s,a_0=a,\pi\Big] \\
&= Q^\pi(s,a).
\end{align}
Thus $Q^\pi$ is a fixed point of $\gT^\nu$.
By uniqueness of the fixed point, $Q^\star = Q^\pi$. Therefore, $\gT^\nu$ converges to $Q^\pi$ under repeated application, i.e., $Q_{n+1}=\gT^\nu Q_n$ implies $Q_n \to Q^\pi$ in $\|\cdot\|_\infty$.
% \end{proof}

\section{Experimental Details}
\label{sec:appendix_exp_details}
\subsection{Tasks}
\label{subsec:tasks}

\paragraph{OGBench.}
All experiments are conducted on OGBench, a large-scale benchmark originally designed for offline goal-conditioned reinforcement learning. Following prior work, we use the single-task variants (``-singletask'') of OGBench to benchmark standard reward-maximizing offline RL methods. Each OGBench environment provides five evaluation goals, corresponding to different tasks, where transitions in the dataset are relabeled with a semi-sparse reward function for a fixed goal. The episode terminates when the agent reaches the goal configuration. We refer readers to prior work for a detailed description of the benchmark and reward construction \cite{ogbench_park2025, fql_park}.

Our main experiments are conducted on three task sets, each designed to evaluate different challenges in offline RL.

\textbf{Standard tasks.}
The first task set consists of 10 standard environments (50 tasks in total) commonly used in prior offline RL studies. This set includes five locomotion environments and five manipulation environments:
\begin{itemize}
    \item \textbf{Locomotion}
    \begin{itemize}
        \item \texttt{antmaze-large-navigate}
        \item \texttt{antmaze-giant-navigate}
        \item \texttt{humanoidmaze-medium-navigate}
        \item \texttt{humanoidmaze-large-navigate}
        \item \texttt{antsoccer-arena-navigate}
    \end{itemize}
    \item \textbf{Manipulation}
    \begin{itemize}
        \item \texttt{cube-single-play}
        \item \texttt{cube-double-play}
        \item \texttt{scene-play}
        \item \texttt{puzzle-3x3-play}
        \item \texttt{puzzle-4x4-play}
    \end{itemize}
\end{itemize}

\textbf{Noisy tasks.}
The second task set focuses on learning from highly suboptimal datasets, where effective policies must be recovered despite limited or noisy coverage. This set includes five environments (25 tasks in total) with exploratory or noisy data distributions:
\begin{itemize}
    \item \texttt{antmaze-large-explore}
    \item \texttt{antmaze-giant-explore}
    \item \texttt{cube-double-play-noisy}
    \item \texttt{puzzle-4x4-play-noisy}
    \item \texttt{scene-play-noisy}
\end{itemize}

\textbf{Long-horizon reasoning tasks.}
The third task set consists of long-horizon reasoning tasks, where reaching the target configuration requires significantly more environment steps compared to the previous task sets. This set includes five environments (25 tasks in total) as follows:
\begin{itemize}
    \item \texttt{cube-triple-play}
    \item \texttt{cube-quadruple-play}
    \item \texttt{puzzle-4x5-play}
    \item \texttt{puzzle-4x6-play}
    \item \texttt{humanoidmaze-giant-navigate}
\end{itemize}
We note that \texttt{cube-octuple-play} is excluded since none of the baselines show meaningful results.

\subsection{Baselines and Hyperparameters}
\label{subsec:hyperparameters}
In this section, we describe the baselines used in our main experiments along with their hyperparameter settings. Hyperparameters that are shared across methods and environments are reported in Table~\ref{tab:common_hyperparam}. For method-specific hyperparameters that require tuning, we follow the protocol of \citet{fql_park} and tune each method on the default task of each environment. Below, we provide a brief description of each baseline and the method-specific hyperparameters that we tuned. For model-based baselines, we use the codebase of \citet{mac_park}.
\begin{itemize}
    \item \textbf{IQL \cite{iql_kostrikov}} trains a critic using expectile regression, which prevents querying out-of-distribution actions when computing TD targets. The policy is trained via advantage-weighted regression. We use the results reported in \citet{fql_park}.
    
    \item \textbf{ReBRAC \cite{rebrac_tarasov}} performs SARSA-style updates while regularizing the training objective with the mean-squared error between behavior actions and actions sampled from the current policy. We use the results reported in \citet{fql_park}.
    
    \item \textbf{FQL \cite{fql_park}} trains a one-step push-forward policy regularized by a behavior flow-matching objective, and updates the critic using SARSA. We use the results reported in \citet{fql_park}.
    
    \item \textbf{MOPO \cite{mopo_yu}} generates synthetic transitions using learned dynamics models and penalizes the rewards of synthetic states based on model uncertainty. For standard manipulation tasks, we use the results reported in \citet{mac_park}. For locomotion tasks, we sweep over penalty coefficients $\omega \in \{0.1, 0.5, 1.0, 2.0, 3.0, 5.0\}$, and fix the rollout length $H=10$ and the model batch ratio $f=0.25$. Since MOPO consistently yields near-zero success rates on locomotion tasks, we do not report individual values.
    
    \item \textbf{MOBILE \cite{mobile_sun}} replaces the dynamics uncertainty penalty in MOPO with uncertainty estimates in the Bellman backup targets. Hyperparameter tuning follows the same protocol as MOPO. However, similar to MOPO, it also achieves zero reward on every locomotion environment, so we do not report the environment-specific values.
    
    \item \textbf{MAC \cite{mac_park}} employs a flow-based action-chunking policy and generates long-horizon rollouts by deploying it over learned dynamics models. For standard manipulation tasks, we use the results reported in \citet{mac_park}. For locomotion tasks, we follow the default hyperparameter settings provided in the paper.
\end{itemize}

We now present an explanation of additional baselines and the method-specific hyperparameters. Our implementation of additional baselines and the proposed method is based on the codebase of \citet{fql_park}.
\begin{itemize}
    \item \textbf{ReBRAC$^\dagger$} is tuned under our experimental setup, using the common hyperparameters reported in Table~\ref{tab:common_hyperparam}. Compared to \citet{fql_park}, we use a larger discount factor $\gamma=0.999$ for all environments. We only tune the actor BC coefficient $\alpha$ separately for each environment, and fix the critic BC coefficient to zero as it has a relatively marginal effect.
    
    \item \textbf{MBTD($\lambda$)} is a model-based approach that generates $n$-step rollouts using a single-step dynamics model and performs TD($\lambda$) on the generated trajectories. For controlled ablation studies, we model the dynamics using flow-matching with the same number of flow steps as our method. We also apply the same techniques used in our approach, including $\lambda$ scheduling, terminal state handling, and horizon winsorization. We only select the behavior-cloning coefficient $\alpha$ separately for each environment.
    
    \item \textbf{DTD($\lambda$)} is a model-free method that performs TD($\lambda$) on trajectories sampled directly from the dataset. As with MBTD($\lambda$), we apply the same techniques as our method, including $\lambda$ scheduling, terminal state handling, and horizon winsorization. We only select the behavior-cloning coefficient $\alpha$ separately for each environment.

    \item \textbf{GHM} is a model-based method that uses geometric horizon models to predict discounted future states and performs $\gamma$-MVE. It is identical to our method except that the horizon $n$ is not provided as an input to the model and horizon winsorization is not applicable. We only select the behavior-cloning coefficient $\alpha$ separately for each environment.

\end{itemize}
To ensure a fair comparison, the proposed method and additional baselines share the same design choices, such as $\lambda$ scheduling, and hyperparameters reported in Table \ref{tab:uhm_hyperparam}. The only hyperparameter we tuned is the actor BC coefficient $\alpha$. It is selected based on the performance of DTD($\lambda$) for each environment from the candidate set $\{0.003, 0.01, 0.03, 0.1, 0.3, 1.0\}$. The selected BC coefficients for each environment are reported in Table \ref{tab_bc_coeff}.

% ReBRAC$^\dagger$ and DTD($\lambda$) share the same coefficient $\alpha$, which is selected based 

% Similarly, GHM, MBTD($\lambda$), and our method share a common coefficient $\alpha$, which is selected based on the performance of UHM from the same candidate set. 
\clearpage
\begin{table}[p]
\centering
\caption{Common hyperparameters for offline RL experiment.}
\label{tab:common_hyperparam}
\small
\begin{tabular}{l l}
\toprule
\textbf{Hyperparameter} & \textbf{Value} \\
\midrule
Dataset size & 1M (default), 10M (long-horizon reasoning tasks) \\
Learning rate & 0.0003 \\
Optimizer & Adam \cite{kingma2014adam} \\
Gradient steps & 1M (default), 2M (long-horizon reasoning tasks) \\
Minibatch size & 256 \\
MLP dimensions & [512, 512, 512, 512] \\
Nonlinearity & GELU \cite{hendrycks2016gaussian} \\
EMA decay $\eta$ & 0.005 \\
\bottomrule
\end{tabular}
\end{table}

\begin{table}[p]
\centering
\caption{Common hyperparameters of the proposed method and additional baselines.}
\label{tab:uhm_hyperparam}
\small
\begin{tabular}{l l}
\toprule
\textbf{Hyperparameter} & \textbf{Value} \\
\midrule
Discount factor $\gamma$ & 0.999 \\
Clipped double Q-learning & True (locomotion tasks), False (manipulation tasks) \\
Flow steps & 5 \\
ODE solver & Midpoint \\
Final td-lambda $\lambda_f$ & 0.8 (default), 0.9 (long-horizon reasoning tasks) \\
Behavior mixing coefficient $\beta$ & 0.3 \\
Winsorization quantile $q$ & 0.2 \\
% BC coefficient $\alpha$ & Tables 6 and 7 \\
\bottomrule
\end{tabular}
\end{table}

\begin{table*}[p]
\caption{BC coefficient $\alpha$ used in main experiments.}
\label{tab_bc_coeff}
\centering
\small
% Resize the table to 1\textwidth (the full width of the text area)
    \begin{tabular}{lccccc}
    \toprule
     & ReBRA$\text{C}^\dagger$ & MBTD($\lambda$) & DTD($\lambda$) & GHM & UHM \\
    \midrule
    antmaze-large-navigate  & 0.01 & 0.01 & 0.01 & 0.01 & 0.01 \\
    antmaze-giant-navigate  & 0.01 & 0.01 & 0.01 & 0.01 & 0.01 \\
    humanoidmaze-medium-navigate & 0.01 & 0.01 & 0.01 & 0.01 & 0.01 \\
    humanoidmaze-large-navigate  & 0.01 & 0.01 & 0.01 & 0.01 & 0.01 \\
    antsoccer-arena-navigate & 0.01 & 0.01 & 0.01 & 0.01 & 0.01 \\
    cube-single-play & 1.0 & 1.0 & 1.0 & 1.0 & 1.0 \\
    cube-double-play  & 0.1 & 0.1 & 0.1 & 0.1 & 0.1 \\
    scene-play  & 0.1 & 0.1 & 0.1 & 0.1 & 0.1 \\
    puzzle-3x3-play  & 0.3 & 0.3 & 0.3 & 0.3 & 0.3 \\
    puzzle-4x4-play  & 0.1 & 0.1 & 0.1 & 0.1 & 0.1 \\
    \midrule
    antmaze-medium-explore  & 0.003 & 0.01 & 0.003 & 0.01 & 0.01 \\
    antmaze-large-explore  & 0.003 & 0.01 & 0.003 & 0.01 & 0.01 \\
    cube-double-noisy  & 0.01 & 0.01 & 0.01 & 0.01 & 0.01 \\
    scene-noisy  & 0.03 & 0.03 & 0.03 & 0.03 & 0.03 \\
    puzzle-4x4-noisy  & 0.01 & 0.01 & 0.01 & 0.01 & 0.01 \\
    \midrule
    cube-triple-play  & 0.1 & 0.1 & 0.1 & 0.1 & 0.1 \\
    cube-quadruple-play  & 0.03 & 0.03 & 0.03 & 0.03 & 0.03 \\
    puzzle-4x5-play  & 0.01 & 0.01 & 0.01 & 0.01 & 0.01 \\
    puzzle-4x6-play  & 0.01 & 0.01 & 0.01 & 0.01 & 0.01 \\
    humanoidmaze-giant-navigate  & 0.01 & 0.01 & 0.01 & 0.01 & 0.01 \\
    \bottomrule
    \end{tabular}
\end{table*}

\clearpage
\section{Additional Results}
\label{sec:appendix_additional_results}
In this section, we provide the full experimental results across 100 OGBench tasks. Table~\ref{ogbench_regular_full_table} reports per-task results on 50 standard tasks, where each result is averaged over five independent seeds and accompanied by the corresponding standard deviation. Table~\ref{ogbench_stitching_full_table} presents per-task results on 25 noisy tasks, again averaged over five independent seeds with standard deviations reported. Finally, Table~\ref{ogbench_long_horizon_full_table} reports per-task results on 25 long-horizon reasoning tasks, where results are averaged over three independent seeds with standard deviations.

\begin{table*}[h]
\caption{Full results on standard tasks in OGBench.}
\label{ogbench_regular_full_table}
\centering
\small
\resizebox{\textwidth}{!}{
    \begin{tabular}{lcccccccccccc}
    \toprule
     & \multicolumn{3}{c}{\textbf{Model-Free}} 
     & \multicolumn{3}{c}{\textbf{Model-Based}}
     & \multicolumn{4}{c}{\textbf{Ablations}}
     & \multicolumn{1}{c}{\textbf{Ours}} \\
    \cmidrule(lr){2-4} \cmidrule(lr){5-7} \cmidrule(lr){8-11} \cmidrule(lr){12-12}
     & IQL & ReBRAC & FQL & MOPO & MOBILE & MAC & ReBRA$\text{C}^\dagger$ & MBTD($\lambda$) & DTD($\lambda$) & GHM & UHM \\
    \midrule
    antmaze-large-navigate-task1  & 48 $\pm$ 9 & 91 $\pm$ 10 & 80 $\pm$ 8 & 0 $\pm$ 0 & 0 $\pm$ 0 & 17 $\pm$ 12 & 67 $\pm$ 41 & 75 $\pm$ 23 & 98 $\pm$ 1 & 96 $\pm$ 1 & 95 $\pm$ 1 \\
    antmaze-large-navigate-task2  & 42 $\pm$ 6 & 88 $\pm$ 4 & 57 $\pm$ 10 & 0 $\pm$ 0 & 0 $\pm$ 0 & 0 $\pm$ 0 & 75 $\pm$ 24 & 68 $\pm$ 28 & 86 $\pm$ 4 & 82 $\pm$ 4 & 80 $\pm$ 5 \\
    antmaze-large-navigate-task3  & 72 $\pm$ 7 & 51 $\pm$ 18 & 93 $\pm$ 3 & 0 $\pm$ 0 & 0 $\pm$ 0 & 70 $\pm$ 14 & 78 $\pm$ 39 & 85 $\pm$ 14 & 99 $\pm$ 1 & 93 $\pm$ 3 & 93 $\pm$ 3 \\
    antmaze-large-navigate-task4  & 51 $\pm$ 9 & 84 $\pm$ 7 & 80 $\pm$ 4 & 0 $\pm$ 0 & 0 $\pm$ 0 & 0 $\pm$ 0 & 63 $\pm$ 35 & 81 $\pm$ 12 & 90 $\pm$ 4 & 91 $\pm$ 2 & 86 $\pm$ 5 \\
    antmaze-large-navigate-task5  & 54 $\pm$ 22 & 90 $\pm$ 2 & 83 $\pm$ 4 & 0 $\pm$ 0 & 0 $\pm$ 0 & 1 $\pm$ 1 & 75 $\pm$ 38 & 45 $\pm$ 37 & 95 $\pm$ 2 & 90 $\pm$ 1 & 91 $\pm$ 3 \\
    \midrule
    antmaze-giant-navigate-task1  & 0 $\pm$ 0 & 27 $\pm$ 22 & 4 $\pm$ 5 & 0 $\pm$ 0 & 0 $\pm$ 0 & 0 $\pm$ 0 & 7 $\pm$ 11 & 11 $\pm$ 14 & 27 $\pm$ 34 & 37 $\pm$ 14 & 48 $\pm$ 5 \\
    antmaze-giant-navigate-task2  & 1 $\pm$ 1 & 16 $\pm$ 17 & 9 $\pm$ 7 & 0 $\pm$ 0 & 0 $\pm$ 0 & 0 $\pm$ 0 & 26 $\pm$ 31 & 40 $\pm$ 17 & 94 $\pm$ 3 & 1 $\pm$ 1 & 5 $\pm$ 6 \\
    antmaze-giant-navigate-task3  & 0 $\pm$ 0 & 34 $\pm$ 22 & 0 $\pm$ 1 & 0 $\pm$ 0 & 0 $\pm$ 0 & 0 $\pm$ 0 & 23 $\pm$ 27 & 14 $\pm$ 12 & 59 $\pm$ 17 & 1 $\pm$ 2 & 7 $\pm$ 14 \\
    antmaze-giant-navigate-task4  & 0 $\pm$ 0 & 5 $\pm$ 12 & 14 $\pm$ 23 & 0 $\pm$ 0 & 0 $\pm$ 0 & 0 $\pm$ 0 & 17 $\pm$ 22 & 2 $\pm$ 2 & 30 $\pm$ 35 & 58 $\pm$ 27 & 60 $\pm$ 27 \\
    antmaze-giant-navigate-task5  & 19 $\pm$ 7 & 49 $\pm$ 22 & 16 $\pm$ 28 & 0 $\pm$ 0 & 0 $\pm$ 0 & 0 $\pm$ 0 & 79 $\pm$ 2 & 67 $\pm$ 17 & 49 $\pm$ 10 & 67 $\pm$ 9 & 59 $\pm$ 14 \\
    \midrule
    humanoidmaze-medium-navigate-task1  & 32 $\pm$ 7 & 16 $\pm$ 9 & 19 $\pm$ 12 & 0 $\pm$ 0 & 0 $\pm$ 0 & 0 $\pm$ 1 & 19 $\pm$ 13 & 34 $\pm$ 9 & 82 $\pm$ 17 & 89 $\pm$ 2 & 95 $\pm$ 1 \\
    humanoidmaze-medium-navigate-task2  & 41 $\pm$ 9 & 18 $\pm$ 16 & 94 $\pm$ 3 & 0 $\pm$ 0 & 0 $\pm$ 0 & 2 $\pm$ 1 & 15 $\pm$ 15 & 78 $\pm$ 35 & 95 $\pm$ 3 & 93 $\pm$ 1 & 96 $\pm$ 1 \\
    humanoidmaze-medium-navigate-task3  & 25 $\pm$ 5 & 36 $\pm$ 13 & 74 $\pm$ 18 & 0 $\pm$ 0 & 0 $\pm$ 0 & 5 $\pm$ 1 & 37 $\pm$ 27 & 93 $\pm$ 2 & 98 $\pm$ 1 & 93 $\pm$ 1 & 94 $\pm$ 2 \\
    humanoidmaze-medium-navigate-task4  & 0 $\pm$ 1 & 15 $\pm$ 16 & 3 $\pm$ 4 & 0 $\pm$ 0 & 0 $\pm$ 0 & 0 $\pm$ 0 & 11 $\pm$ 19 & 19 $\pm$ 23 & 31 $\pm$ 22 & 84 $\pm$ 2 & 92 $\pm$ 4 \\
    humanoidmaze-medium-navigate-task5  & 66 $\pm$ 4 & 24 $\pm$ 20 & 97 $\pm$ 2 & 0 $\pm$ 0 & 0 $\pm$ 0 & 2 $\pm$ 1 & 26 $\pm$ 22 & 97 $\pm$ 1 & 98 $\pm$ 0 & 92 $\pm$ 2 & 96 $\pm$ 1 \\
    \midrule
    humanoidmaze-large-navigate-task1  & 3 $\pm$ 1 & 2 $\pm$ 1 & 7 $\pm$ 6 & 0 $\pm$ 0 & 0 $\pm$ 0 & 0 $\pm$ 0 & 1 $\pm$ 1 & 25 $\pm$ 15 & 51 $\pm$ 22 & 35 $\pm$ 9 & 69 $\pm$ 3 \\
    humanoidmaze-large-navigate-task2  & 0 $\pm$ 0 & 0 $\pm$ 0 & 0 $\pm$ 0 & 0 $\pm$ 0 & 0 $\pm$ 0 & 0 $\pm$ 0 & 0 $\pm$ 0 & 0 $\pm$ 1 & 4 $\pm$ 5 & 1 $\pm$ 1 & 3 $\pm$ 1 \\
    humanoidmaze-large-navigate-task3  & 7 $\pm$ 3 & 8 $\pm$ 4 & 11 $\pm$ 7 & 0 $\pm$ 0 & 0 $\pm$ 0 & 0 $\pm$ 0 & 3 $\pm$ 4 & 31 $\pm$ 10 & 55 $\pm$ 45 & 40 $\pm$ 2 & 56 $\pm$ 17 \\
    humanoidmaze-large-navigate-task4  & 1 $\pm$ 0 & 1 $\pm$ 1 & 2 $\pm$ 3 & 0 $\pm$ 0 & 0 $\pm$ 0 & 0 $\pm$ 0 & 0 $\pm$ 0 & 10 $\pm$ 6 & 8 $\pm$ 12 & 2 $\pm$ 1 & 17 $\pm$ 22 \\
    humanoidmaze-large-navigate-task5  & 1 $\pm$ 1 & 2 $\pm$ 2 & 1 $\pm$ 3 & 0 $\pm$ 0 & 0 $\pm$ 0 & 0 $\pm$ 0 & 0 $\pm$ 1 & 12 $\pm$ 10 & 19 $\pm$ 25 & 3 $\pm$ 5 & 17 $\pm$ 14 \\
    \midrule
    antsoccer-arena-navigate-task1  & 14 $\pm$ 5 & 0 $\pm$ 0 & 77 $\pm$ 4 & 0 $\pm$ 0 & 0 $\pm$ 0 & 47 $\pm$ 6 & 0 $\pm$ 0 & 53 $\pm$ 6 & 0 $\pm$ 0 & 14 $\pm$ 4 & 17 $\pm$ 6 \\
    antsoccer-arena-navigate-task2  & 17 $\pm$ 7 & 0 $\pm$ 1 & 88 $\pm$ 3 & 0 $\pm$ 0 & 0 $\pm$ 0 & 30 $\pm$ 10 & 2 $\pm$ 4 & 66 $\pm$ 5 & 0 $\pm$ 0 & 40 $\pm$ 9 & 50 $\pm$ 13 \\
    antsoccer-arena-navigate-task3  & 6 $\pm$ 4 & 0 $\pm$ 0 & 61 $\pm$ 6 & 0 $\pm$ 0 & 0 $\pm$ 0 & 30 $\pm$ 2 & 0 $\pm$ 0 & 56 $\pm$ 4 & 0 $\pm$ 0 & 1 $\pm$ 1 & 8 $\pm$ 12 \\
    antsoccer-arena-navigate-task4  & 3 $\pm$ 2 & 0 $\pm$ 0 & 39 $\pm$ 6 & 0 $\pm$ 0 & 0 $\pm$ 0 & 25 $\pm$ 1 & 1 $\pm$ 1 & 31 $\pm$ 3 & 1 $\pm$ 1 & 23 $\pm$ 4 & 25 $\pm$ 4 \\
    antsoccer-arena-navigate-task5  & 2 $\pm$ 2 & 0 $\pm$ 0 & 36 $\pm$ 9 & 0 $\pm$ 0 & 0 $\pm$ 0 & 15 $\pm$ 8 & 0 $\pm$ 1 & 27 $\pm$ 10 & 0 $\pm$ 0 & 22 $\pm$ 14 & 28 $\pm$ 7 \\
    \midrule
    cube-single-play-task1  & 88 $\pm$ 3 & 89 $\pm$ 5 & 97 $\pm$ 2 & 12 $\pm$ 16 & 85 $\pm$ 22 & 100 $\pm$ 0 & 93 $\pm$ 3 & 91 $\pm$ 2 & 91 $\pm$ 7 & 87 $\pm$ 6 & 95 $\pm$ 6 \\
    cube-single-play-task2  & 85 $\pm$ 8 & 92 $\pm$ 4 & 97 $\pm$ 2 & 10 $\pm$ 16 & 80 $\pm$ 12 & 100 $\pm$ 0 & 89 $\pm$ 3 & 93 $\pm$ 3 & 91 $\pm$ 4 & 93 $\pm$ 4 & 94 $\pm$ 6 \\
    cube-single-play-task3  & 91 $\pm$ 5 & 93 $\pm$ 3 & 98 $\pm$ 2 & 15 $\pm$ 14 & 83 $\pm$ 17 & 98 $\pm$ 3 & 94 $\pm$ 3 & 96 $\pm$ 2 & 90 $\pm$ 6 & 93 $\pm$ 4 & 93 $\pm$ 5 \\
    cube-single-play-task4  & 73 $\pm$ 6 & 92 $\pm$ 3 & 94 $\pm$ 3 & 2 $\pm$ 3 & 72 $\pm$ 19 & 98 $\pm$ 3 & 92 $\pm$ 5 & 92 $\pm$ 2 & 92 $\pm$ 5 & 92 $\pm$ 2 & 88 $\pm$ 6 \\
    cube-single-play-task5  & 78 $\pm$ 9 & 87 $\pm$ 8 & 93 $\pm$ 3 & 20 $\pm$ 26 & 87 $\pm$ 19 & 97 $\pm$ 7 & 89 $\pm$ 4 & 86 $\pm$ 4 & 87 $\pm$ 6 & 90 $\pm$ 5 & 89 $\pm$ 2 \\
    \midrule
    cube-double-play-task1  & 27 $\pm$ 5 & 45 $\pm$ 6 & 61 $\pm$ 9 & 2 $\pm$ 3 & 7 $\pm$ 8 & 82 $\pm$ 15 & 12 $\pm$ 6 & 12 $\pm$ 4 & 3 $\pm$ 2 & 42 $\pm$ 7 & 44 $\pm$ 10 \\
    cube-double-play-task2  & 1 $\pm$ 1 & 7 $\pm$ 3 & 36 $\pm$ 6 & 0 $\pm$ 0 & 0 $\pm$ 0 & 50 $\pm$ 12 & 6 $\pm$ 3 & 3 $\pm$ 1 & 2 $\pm$ 1 & 34 $\pm$ 7 & 34 $\pm$ 11 \\
    cube-double-play-task3  & 0 $\pm$ 0 & 4 $\pm$ 1 & 22 $\pm$ 5 & 2 $\pm$ 3 & 0 $\pm$ 0 & 55 $\pm$ 10 & 2 $\pm$ 1 & 1 $\pm$ 1 & 0 $\pm$ 1 & 25 $\pm$ 6 & 26 $\pm$ 7 \\
    cube-double-play-task4  & 0 $\pm$ 0 & 1 $\pm$ 1 & 5 $\pm$ 2 & 0 $\pm$ 0 & 0 $\pm$ 0 & 28 $\pm$ 8 & 0 $\pm$ 0 & 0 $\pm$ 0 & 0 $\pm$ 0 & 1 $\pm$ 1 & 1 $\pm$ 0 \\
    cube-double-play-task5  & 4 $\pm$ 3 & 4 $\pm$ 2 & 19 $\pm$ 10 & 2 $\pm$ 3 & 0 $\pm$ 0 & 50 $\pm$ 9 & 1 $\pm$ 1 & 1 $\pm$ 1 & 14 $\pm$ 3 & 41 $\pm$ 2 & 43 $\pm$ 9 \\
    \midrule
    scene-play-task1  & 94 $\pm$ 3 & 95 $\pm$ 2 & 100 $\pm$ 0 & 30 $\pm$ 38 & 37 $\pm$ 16 & 100 $\pm$ 0 & 82 $\pm$ 8 & 68 $\pm$ 14 & 97 $\pm$ 3 & 89 $\pm$ 5 & 84 $\pm$ 4 \\
    scene-play-task2  & 12 $\pm$ 3 & 50 $\pm$ 13 & 76 $\pm$ 9 & 2 $\pm$ 3 & 5 $\pm$ 10 & 100 $\pm$ 0 & 65 $\pm$ 5 & 59 $\pm$ 4 & 96 $\pm$ 2 & 89 $\pm$ 4 & 87 $\pm$ 2 \\
    scene-play-task3  & 32 $\pm$ 7 & 55 $\pm$ 16 & 98 $\pm$ 1 & 0 $\pm$ 0 & 0 $\pm$ 0 & 95 $\pm$ 10 & 44 $\pm$ 11 & 26 $\pm$ 11 & 48 $\pm$ 12 & 24 $\pm$ 8 & 23 $\pm$ 9 \\
    scene-play-task4  & 0 $\pm$ 1 & 3 $\pm$ 3 & 5 $\pm$ 1 & 0 $\pm$ 0 & 0 $\pm$ 0 & 95 $\pm$ 6 & 9 $\pm$ 14 & 1 $\pm$ 1 & 75 $\pm$ 19 & 3 $\pm$ 3 & 4 $\pm$ 6 \\
    scene-play-task5  & 0 $\pm$ 0 & 0 $\pm$ 0 & 0 $\pm$ 0 & 0 $\pm$ 0 & 0 $\pm$ 0 & 93 $\pm$ 8 & 0 $\pm$ 0 & 0 $\pm$ 0 & 63 $\pm$ 9 & 15 $\pm$ 19 & 19 $\pm$ 27 \\
    \midrule
    puzzle-3x3-play-task1  & 33 $\pm$ 6 & 97 $\pm$ 4 & 90 $\pm$ 4 & 100 $\pm$ 0 & 60 $\pm$ 47 & 100 $\pm$ 0 & 96 $\pm$ 4 & 98 $\pm$ 1 & 100 $\pm$ 0 & 93 $\pm$ 3 & 99 $\pm$ 1 \\
    puzzle-3x3-play-task2  & 4 $\pm$ 3 & 1 $\pm$ 1 & 16 $\pm$ 5 & 0 $\pm$ 0 & 0 $\pm$ 0 & 0 $\pm$ 0 & 89 $\pm$ 8 & 94 $\pm$ 3 & 100 $\pm$ 0 & 49 $\pm$ 2 & 100 $\pm$ 0 \\
    puzzle-3x3-play-task3  & 3 $\pm$ 2 & 3 $\pm$ 1 & 10 $\pm$ 3 & 0 $\pm$ 0 & 0 $\pm$ 0 & 0 $\pm$ 0 & 83 $\pm$ 10 & 85 $\pm$ 5 & 99 $\pm$ 1 & 35 $\pm$ 7 & 99 $\pm$ 1 \\
    puzzle-3x3-play-task4  & 2 $\pm$ 1 & 2 $\pm$ 1 & 16 $\pm$ 5 & 0 $\pm$ 0 & 0 $\pm$ 0 & 0 $\pm$ 0 & 89 $\pm$ 6 & 93 $\pm$ 3 & 100 $\pm$ 0 & 35 $\pm$ 4 & 99 $\pm$ 1 \\
    puzzle-3x3-play-task5  & 3 $\pm$ 2 & 5 $\pm$ 3 & 16 $\pm$ 3 & 0 $\pm$ 0 & 0 $\pm$ 0 & 0 $\pm$ 0 & 92 $\pm$ 1 & 94 $\pm$ 3 & 99 $\pm$ 2 & 44 $\pm$ 7 & 98 $\pm$ 1 \\
    \midrule
    puzzle-4x4-play-task1  & 12 $\pm$ 2 & 26 $\pm$ 4 & 34 $\pm$ 8 & 0 $\pm$ 0 & 0 $\pm$ 0 & 98 $\pm$ 3 & 2 $\pm$ 1 & 9 $\pm$ 3 & 1 $\pm$ 1 & 24 $\pm$ 4 & 19 $\pm$ 6 \\
    puzzle-4x4-play-task2  & 7 $\pm$ 4 & 12 $\pm$ 4 & 16 $\pm$ 5 & 0 $\pm$ 0 & 0 $\pm$ 0 & 33 $\pm$ 27 & 1 $\pm$ 0 & 1 $\pm$ 1 & 1 $\pm$ 0 & 11 $\pm$ 1 & 11 $\pm$ 5 \\
    puzzle-4x4-play-task3  & 9 $\pm$ 3 & 15 $\pm$ 3 & 18 $\pm$ 5 & 0 $\pm$ 0 & 0 $\pm$ 0 & 100 $\pm$ 0 & 1 $\pm$ 0 & 6 $\pm$ 2 & 1 $\pm$ 0 & 14 $\pm$ 2 & 13 $\pm$ 4 \\
    puzzle-4x4-play-task4  & 5 $\pm$ 2 & 10 $\pm$ 3 & 11 $\pm$ 3 & 0 $\pm$ 0 & 0 $\pm$ 0 & 85 $\pm$ 14 & 2 $\pm$ 1 & 3 $\pm$ 1 & 0 $\pm$ 1 & 8 $\pm$ 2 & 8 $\pm$ 2 \\
    puzzle-4x4-play-task5  & 4 $\pm$ 1 & 7 $\pm$ 3 & 7 $\pm$ 3 & 0 $\pm$ 0 & 0 $\pm$ 0 & 72 $\pm$ 40 & 1 $\pm$ 1 & 1 $\pm$ 1 & 0 $\pm$ 1 & 8 $\pm$ 3 & 7 $\pm$ 2 \\
    \midrule
    Average  & 23 & 31 & 44 & 4 & 10 & 40 & 35 $\pm$ 2 & 45 $\pm$ 2 & 52 $\pm$ 1 & 48 $\pm$ 1 & 55 $\pm$ 1 \\
    \bottomrule
    \end{tabular}
}
\end{table*}

\begin{table*}[h]
\caption{Full results on noisy tasks in OGBench.}
\label{ogbench_stitching_full_table}
\centering
\small
% Resize the table to 1\textwidth (the full width of the text area)
% \resizebox{\textwidth}{!}{
    \begin{tabular}{lccccc}
    \toprule
     & ReBRA$\text{C}^\dagger$ & MBTD($\lambda$) & DTD($\lambda$) & GHM & UHM \\
    \midrule
    antmaze-medium-explore-task1  & 38 $\pm$ 16 & 98 $\pm$ 2 & 55 $\pm$ 25 & 97 $\pm$ 1 & 95 $\pm$ 6 \\
    antmaze-medium-explore-task2  & 97 $\pm$ 3 & 99 $\pm$ 1 & 99 $\pm$ 1 & 96 $\pm$ 2 & 97 $\pm$ 1 \\
    antmaze-medium-explore-task3  & 64 $\pm$ 16 & 86 $\pm$ 13 & 75 $\pm$ 11 & 84 $\pm$ 12 & 70 $\pm$ 10 \\
    antmaze-medium-explore-task4  & 72 $\pm$ 15 & 98 $\pm$ 1 & 76 $\pm$ 14 & 84 $\pm$ 6 & 89 $\pm$ 6 \\
    antmaze-medium-explore-task5  & 98 $\pm$ 1 & 100 $\pm$ 0 & 100 $\pm$ 1 & 94 $\pm$ 2 & 95 $\pm$ 6 \\
    \midrule
    antmaze-large-explore-task1  & 3 $\pm$ 5 & 21 $\pm$ 13 & 11 $\pm$ 11 & 22 $\pm$ 9 & 56 $\pm$ 16 \\
    antmaze-large-explore-task2  & 15 $\pm$ 12 & 0 $\pm$ 0 & 0 $\pm$ 0 & 0 $\pm$ 0 & 2 $\pm$ 3 \\
    antmaze-large-explore-task3  & 40 $\pm$ 33 & 79 $\pm$ 12 & 57 $\pm$ 40 & 48 $\pm$ 12 & 68 $\pm$ 10 \\
    antmaze-large-explore-task4  & 5 $\pm$ 5 & 0 $\pm$ 0 & 13 $\pm$ 12 & 0 $\pm$ 1 & 0 $\pm$ 0 \\
    antmaze-large-explore-task5  & 1 $\pm$ 1 & 0 $\pm$ 0 & 8 $\pm$ 8 & 0 $\pm$ 1 & 6 $\pm$ 5 \\
    \midrule
    cube-double-noisy-task1  & 10 $\pm$ 3 & 9 $\pm$ 2 & 10 $\pm$ 7 & 59 $\pm$ 11 & 52 $\pm$ 6 \\
    cube-double-noisy-task2  & 1 $\pm$ 1 & 0 $\pm$ 1 & 1 $\pm$ 0 & 28 $\pm$ 10 & 15 $\pm$ 6 \\
    cube-double-noisy-task3  & 0 $\pm$ 0 & 1 $\pm$ 1 & 0 $\pm$ 0 & 11 $\pm$ 3 & 7 $\pm$ 1 \\
    cube-double-noisy-task4  & 0 $\pm$ 0 & 0 $\pm$ 0 & 0 $\pm$ 0 & 12 $\pm$ 4 & 9 $\pm$ 2 \\
    cube-double-noisy-task5  & 1 $\pm$ 1 & 0 $\pm$ 0 & 0 $\pm$ 0 & 10 $\pm$ 3 & 8 $\pm$ 5 \\
    \midrule
    scene-noisy-task1  & 93 $\pm$ 3 & 91 $\pm$ 2 & 26 $\pm$ 18 & 98 $\pm$ 1 & 96 $\pm$ 2 \\
    scene-noisy-task2  & 0 $\pm$ 0 & 22 $\pm$ 5 & 0 $\pm$ 0 & 63 $\pm$ 8 & 69 $\pm$ 7 \\
    scene-noisy-task3  & 33 $\pm$ 32 & 70 $\pm$ 7 & 5 $\pm$ 4 & 82 $\pm$ 6 & 83 $\pm$ 9 \\
    scene-noisy-task4  & 9 $\pm$ 8 & 28 $\pm$ 9 & 0 $\pm$ 0 & 41 $\pm$ 9 & 54 $\pm$ 13 \\
    scene-noisy-task5  & 0 $\pm$ 0 & 0 $\pm$ 0 & 0 $\pm$ 0 & 0 $\pm$ 0 & 2 $\pm$ 3 \\
    \midrule
    puzzle-4x4-noisy-task1  & 0 $\pm$ 0 & 1 $\pm$ 1 & 0 $\pm$ 0 & 18 $\pm$ 3 & 2 $\pm$ 1 \\
    puzzle-4x4-noisy-task2  & 0 $\pm$ 0 & 0 $\pm$ 0 & 0 $\pm$ 0 & 0 $\pm$ 0 & 0 $\pm$ 0 \\
    puzzle-4x4-noisy-task3  & 0 $\pm$ 0 & 0 $\pm$ 0 & 0 $\pm$ 0 & 3 $\pm$ 2 & 1 $\pm$ 1 \\
    puzzle-4x4-noisy-task4  & 0 $\pm$ 0 & 0 $\pm$ 0 & 0 $\pm$ 0 & 0 $\pm$ 0 & 0 $\pm$ 0 \\
    puzzle-4x4-noisy-task5  & 0 $\pm$ 0 & 0 $\pm$ 0 & 0 $\pm$ 0 & 0 $\pm$ 0 & 0 $\pm$ 0 \\
    \midrule
    Average  & 23 $\pm$ 2 & 32 $\pm$ 1 & 23 $\pm$ 4 & 38 $\pm$ 1 & 39 $\pm$ 1 \\
    \bottomrule
    \end{tabular}
% }
\end{table*}

\begin{table*}[h]
\caption{Full results on long horizon tasks in OGBench.}
\label{ogbench_long_horizon_full_table}
\centering
\small
% Resize the table to 1\textwidth (the full width of the text area)
% \resizebox{\textwidth}{!}{
    \begin{tabular}{lccccc}
    \toprule
     & ReBRA$\text{C}^\dagger$ & MBTD($\lambda$) & DTD($\lambda$) & GHM & UHM \\
    \midrule
    cube-triple-play-task1  & 22 $\pm$ 6 & 52 $\pm$ 13 & 3 $\pm$ 2 & 80 $\pm$ 14 & 91 $\pm$ 5 \\
    cube-triple-play-task2  & 0 $\pm$ 0 & 2 $\pm$ 3 & 0 $\pm$ 0 & 62 $\pm$ 15 & 94 $\pm$ 2 \\
    cube-triple-play-task3  & 0 $\pm$ 0 & 8 $\pm$ 6 & 0 $\pm$ 0 & 61 $\pm$ 14 & 75 $\pm$ 6 \\
    cube-triple-play-task4  & 0 $\pm$ 0 & 0 $\pm$ 0 & 1 $\pm$ 1 & 16 $\pm$ 7 & 21 $\pm$ 7 \\
    cube-triple-play-task5  & 0 $\pm$ 0 & 0 $\pm$ 0 & 0 $\pm$ 0 & 0 $\pm$ 0 & 0 $\pm$ 0 \\
    \midrule
    cube-quadruple-play-task1  & 0 $\pm$ 0 & 0 $\pm$ 0 & 0 $\pm$ 0 & 25 $\pm$ 9 & 34 $\pm$ 14 \\
    cube-quadruple-play-task2  & 0 $\pm$ 0 & 0 $\pm$ 0 & 2 $\pm$ 3 & 28 $\pm$ 21 & 27 $\pm$ 20 \\
    cube-quadruple-play-task3  & 0 $\pm$ 0 & 0 $\pm$ 0 & 0 $\pm$ 0 & 4 $\pm$ 3 & 13 $\pm$ 6 \\
    cube-quadruple-play-task4  & 0 $\pm$ 0 & 0 $\pm$ 0 & 0 $\pm$ 0 & 2 $\pm$ 1 & 4 $\pm$ 4 \\
    cube-quadruple-play-task5  & 0 $\pm$ 0 & 0 $\pm$ 0 & 0 $\pm$ 0 & 0 $\pm$ 0 & 0 $\pm$ 0 \\
    \midrule
    puzzle-4x5-play-task1  & 36 $\pm$ 12 & 23 $\pm$ 14 & 68 $\pm$ 14 & 83 $\pm$ 6 & 80 $\pm$ 2 \\
    puzzle-4x5-play-task2  & 0 $\pm$ 0 & 0 $\pm$ 0 & 0 $\pm$ 0 & 0 $\pm$ 0 & 0 $\pm$ 0 \\
    puzzle-4x5-play-task3  & 0 $\pm$ 0 & 0 $\pm$ 0 & 0 $\pm$ 0 & 0 $\pm$ 0 & 0 $\pm$ 0 \\
    puzzle-4x5-play-task4  & 0 $\pm$ 0 & 0 $\pm$ 0 & 0 $\pm$ 0 & 0 $\pm$ 0 & 0 $\pm$ 0 \\
    puzzle-4x5-play-task5  & 0 $\pm$ 0 & 0 $\pm$ 0 & 0 $\pm$ 0 & 0 $\pm$ 0 & 0 $\pm$ 0 \\
    \midrule
    puzzle-4x6-play-task1  & 55 $\pm$ 24 & 5 $\pm$ 3 & 26 $\pm$ 28 & 26 $\pm$ 17 & 55 $\pm$ 7 \\
    puzzle-4x6-play-task2  & 0 $\pm$ 0 & 0 $\pm$ 0 & 0 $\pm$ 0 & 0 $\pm$ 0 & 0 $\pm$ 0 \\
    puzzle-4x6-play-task3  & 0 $\pm$ 0 & 0 $\pm$ 0 & 0 $\pm$ 0 & 0 $\pm$ 0 & 0 $\pm$ 0 \\
    puzzle-4x6-play-task4  & 0 $\pm$ 0 & 0 $\pm$ 0 & 0 $\pm$ 0 & 0 $\pm$ 0 & 0 $\pm$ 0 \\
    puzzle-4x6-play-task5  & 0 $\pm$ 0 & 0 $\pm$ 0 & 0 $\pm$ 0 & 0 $\pm$ 0 & 0 $\pm$ 0 \\
    \midrule
    humanoidmaze-giant-navigate-task1  & 1 $\pm$ 1 & 0 $\pm$ 0 & 19 $\pm$ 8 & 0 $\pm$ 0 & 3 $\pm$ 1 \\
    humanoidmaze-giant-navigate-task2  & 6 $\pm$ 5 & 1 $\pm$ 1 & 58 $\pm$ 10 & 7 $\pm$ 1 & 13 $\pm$ 5 \\
    humanoidmaze-giant-navigate-task3  & 0 $\pm$ 0 & 1 $\pm$ 1 & 4 $\pm$ 2 & 0 $\pm$ 0 & 0 $\pm$ 0 \\
    humanoidmaze-giant-navigate-task4  & 1 $\pm$ 1 & 2 $\pm$ 1 & 55 $\pm$ 7 & 1 $\pm$ 1 & 1 $\pm$ 1 \\
    humanoidmaze-giant-navigate-task5  & 4 $\pm$ 4 & 21 $\pm$ 5 & 94 $\pm$ 1 & 11 $\pm$ 2 & 33 $\pm$ 2 \\
    \midrule
    Average  & 5 $\pm$ 1 & 5 $\pm$ 1 & 13 $\pm$ 3 & 16 $\pm$ 2 & 22 $\pm$ 1 \\
    \bottomrule
    \end{tabular}
% }
\end{table*}

% You can have as much text here as you want. The main body must be at most $8$
% pages long. For the final version, one more page can be added. If you want, you
% can use an appendix like this one.

% The $\mathtt{\backslash onecolumn}$ command above can be kept in place if you
% prefer a one-column appendix, or can be removed if you prefer a two-column
% appendix.  Apart from this possible change, the style (font size, spacing,
% margins, page numbering, etc.) should be kept the same as the main body.
%%%%%%%%%%%%%%%%%%%%%%%%%%%%%%%%%%%%%%%%%%%%%%%%%%%%%%%%%%%%%%%%%%%%%%%%%%%%%%%
%%%%%%%%%%%%%%%%%%%%%%%%%%%%%%%%%%%%%%%%%%%%%%%%%%%%%%%%%%%%%%%%%%%%%%%%%%%%%%%

\end{document}